\documentclass[preprint,12pt,times]{elsarticle}

\biboptions{numbers}
\usepackage[a4paper,margin=2.5cm]{geometry}

\usepackage{amsmath}
\usepackage{amssymb}
\usepackage{graphicx}
\usepackage{microtype}
\usepackage{xcolor}
\usepackage{booktabs}
\usepackage{multirow}
\usepackage{makecell}
\usepackage{array}
\usepackage{setspace}
\usepackage{float}
\usepackage{placeins}
\usepackage{caption}
\usepackage{tikz}
\usepackage{pgfplots}
\pgfplotsset{compat=1.18}
\usepgfplotslibrary{groupplots}

\usepackage{url}
\usepackage[unicode,hidelinks]{hyperref}

\makeatletter
\def\ps@pprintTitle{%
  \let\@oddhead\@empty
  \let\@evenhead\@empty
  \def\@oddfoot{\hfil\thepage\hfil}%
  \let\@evenfoot\@oddfoot
}
\makeatother
\hypersetup{
  pdftitle={MonoPRIO: Adaptive Prior Conditioning for Unified Monocular 3D Object Detection},
  pdfauthor={Leon Davies, Qinggang Meng, Mohamad Saada, Baihua Li, Simon Sølvsten},
  pdfsubject={Author-formatted research preprint},
  pdfkeywords={Monocular 3D object detection, size estimation, uncertainty estimation, prior models, KITTI}
}

\begin{document}

\begin{frontmatter}

\title{MonoPRIO: Adaptive Prior Conditioning for Unified Monocular 3D Object Detection}

\author[lu]{Leon Davies\corref{cor1}}
\ead{l.davies2@lboro.ac.uk}

\author[lu]{Qinggang Meng}
\ead{q.meng@lboro.ac.uk}

\author[lu]{Mohamad Saada}
\ead{m.saada@lboro.ac.uk}

\author[lu]{Baihua Li}
\ead{b.li@lboro.ac.uk}

\author[sdu]{Simon S{\o}lvsten}
\ead{simos@sam.sdu.dk}

\cortext[cor1]{Corresponding author}

\affiliation[lu]{
  organization={Department of Computer Science, Loughborough University},
  addressline={Epinal Way},
  city={Loughborough},
  postcode={LE11 3TU},
  state={Leicestershire},
  country={United Kingdom}
}

\affiliation[sdu]{
  organization={European Center for Risk \& Resilience Studies, University of Southern Denmark},
  addressline={Degnevej 14},
  city={Esbjerg},
  postcode={6705},
  country={Denmark}
}

\begin{abstract}
Monocular 3D object detection remains challenging because size and depth are underdetermined by single-view evidence, particularly under occlusion, truncation, and scale-depth ambiguity. Although recent methods improve depth and geometric reasoning, metric size remains unstable in unified multi-class settings, where class variability and limited visibility broaden plausible size modes. We propose MonoPRIO, a unified monocular detector that targets this bottleneck through adaptive prior conditioning in the size pathway. MonoPRIO constructs class-aware size prototypes offline, routes each decoder query to a soft mixture prior, applies uncertainty-aware log-space conditioning, and uses Cluster-Aligned Prior (CAP) regularisation on matched positives. On the KITTI test server, among the compared methods reporting all three categories, MonoPRIO achieves the highest Moderate AP for Car, Pedestrian, and Cyclist. In the car-only setting, it obtains the best 3D bounding-box AP across Easy, Moderate, and Hard among the compared methods without additional depth or LiDAR supervision, while requiring lower latency than MonoCLUE under matched measurement conditions. The frozen CLIP RN50 encoder is executed only during offline prior-bank construction and not during detector training or inference. Ablations show that simple prior baselines do not recover the full-model gain and that learned soft routing outperforms uniform and hard routing over the same visual-size bank. Routed injection provides the main improvement; CAP is complementary. MonoPRIO improves over MonoDGP on the nuScenes frontal protocol. The clearest benefits occur in low-data and partially occluded regimes, while atypical geometry or severe visibility loss can cause prior mismatch.
\end{abstract}

% \begin{highlights}
% \item Strong official KITTI unified and car-only 3D AP with camera-only inference.
% \item Stronger accuracy-efficiency trade-off than previous leading methods.
% \item Query-routed priors improve size robustness under monocular ambiguity.
% \item Gains are strongest in unified, low-data, and occluded regimes.
% \end{highlights}


\end{frontmatter}

\section{Introduction}
\begin{figure}[t]
  \centering
  \includegraphics[width=\textwidth]{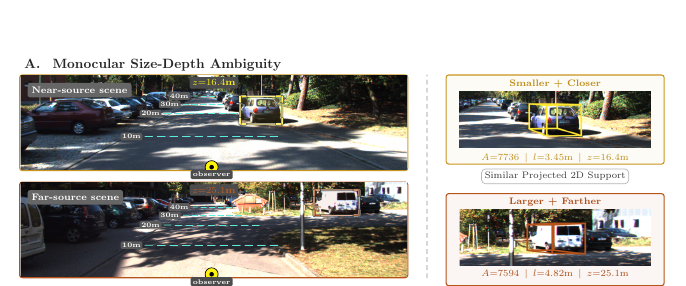}
  \caption{\textbf{Monocular size-depth ambiguity in KITTI scenes.} The highlighted smaller-closer and larger-farther targets exhibit similar projected 2D support in the image, but different metric attributes: the projected support area is similar ($A=7736$ vs.\ $A=7594$), while object length and depth differ ($l=3.45\,\mathrm{m},\, z=16.4\,\mathrm{m}$ vs.\ $l=4.82\,\mathrm{m},\, z=25.1\,\mathrm{m}$). This motivates adaptive prior conditioning for size estimation.}
  \label{fig:intro_teaser}
\end{figure}

Monocular 3D object detection is a central visual perception problem for autonomous driving, as it offers a low-cost camera-only deployment path while remaining fundamentally ill-posed under single-view geometry \cite{Qian_2022_PR_Survey}. Its practical value is clear, but metric 3D recovery from a single image remains challenging, and multiple size-depth configurations can produce highly similar image evidence. This ambiguity becomes more severe under occlusion, truncation, and long-range observations, where visible object extent is limited and geometric cues are weak. Recent methods have advanced monocular 3D detection through stronger depth reasoning, geometry-aware modelling, and query-based transformer decoding \cite{zhang2023monodetr,pu2025monodgp,yang2025monoclue,zhang2025monocop}. Earlier anchor- and centre-guided pipelines also established strong and competitive baselines \cite{lu2021gupnet,li2022monodde,liu2022monocon,jia2023monouni}. However, improved depth prediction and richer query interaction do not fully resolve a persistent mechanism-level issue. Metric size remains unstable when visual evidence is partial. In many difficult scenes, detectors can still fit multiple plausible size hypotheses that are compatible with similar 2D evidence, as illustrated in Fig.~\ref{fig:intro_teaser}. As a result, local depth improvements do not always translate into stable object dimensions, and residual size errors continue to propagate to 3D IoU and bird's-eye-view (BEV) alignment.

This unresolved issue is amplified in unified multi-class settings. A single detector must simultaneously model categories with long-tailed frequency and different intra-class size variability. Under partial visibility, the set of plausible dimensions broadens further, especially for smaller or less frequent classes. Under ambiguous observations, detectors can produce high-variance dimensions, while overly rigid priors can suppress atypical but valid instances. Unified models therefore require a size-estimation strategy that improves stability while preserving instance-specific flexibility.

These observations indicate that an important remaining bottleneck lies in metric size prediction. We therefore focus on branch-local size stabilisation while keeping matching and the remaining prediction heads unchanged. This focus distinguishes MonoPRIO from prior or geometry-informed monocular and camera-only detectors that mainly use priors for depth estimation, anchor or template constraints, image-to-BEV lifting, or broader contextual feature modelling. In contrast, MonoPRIO treats metric size as the direct target of prior conditioning. It does not introduce a new detector backbone, global scene module, or BEV representation, but instead augments the size pathway with query-adaptive class-aware prior routing, uncertainty-modulated log-space size injection, and matched-positive prior regularisation. The contribution is deliberately local. The aim is to stabilise metric size when single-view evidence admits multiple plausible size-depth interpretations, while preserving the base detector's matching and non-size prediction heads.

We instantiate this strategy as MonoPRIO, a branch-local extension to a MonoDGP-style unified monocular 3D detector. MonoPRIO builds class-aware size prototype banks offline, routes each decoder query to a soft mixture prior, conditions log-space size prediction with routed uncertainty, and introduces Cluster-Aligned Prior (CAP), a training-time regularisation term that encourages matched predictions to remain close to plausible prototype manifolds. Frozen CLIP RN50 features~\cite{radford2021clip} are used only during offline bank construction, and the CLIP encoder is not executed during detector training or inference. This preserves the base detector pipeline and keeps the additional inference cost close to the MonoDGP base.

On the official KITTI test server, among the compared methods reporting all three categories, the unified MonoPRIO model achieves the highest Moderate AP for Car, Pedestrian, and Cyclist while maintaining near-MonoDGP efficiency. On the car-only benchmark, MonoPRIO also obtains the best 3D bounding-box AP across Easy/Moderate/Hard on the official KITTI test set among the compared methods using the standard KITTI detection annotations without additional depth or LiDAR supervision. To test whether the prior-conditioning mechanism remains useful outside the primary KITTI evaluation, we further evaluate on the nuScenes frontal protocol under the same KITTI-style conversion setting. MonoPRIO improves over MonoDGP across all reported nuScenes metrics, although MonoCoP remains stronger. We provide mechanism ablations and diagnostics across ambiguity regimes, routing behaviour, low-data settings, additional dataset validation, and quantitative failure analysis.

Our main contributions are summarised as follows

\begin{itemize}
  \item We propose MonoPRIO, a branch-local adaptive size-prior conditioning mechanism for unified monocular 3D detection. MonoPRIO augments the size pathway of a MonoDGP-style detector while keeping matching and the remaining prediction heads unchanged.

  \item We introduce an offline class-aware prototype bank and query-level soft routing mechanism that produces instance-specific size priors, reducing the rigidity of fixed class templates while preserving class-consistent support.

  \item We formulate uncertainty-aware log-space size injection and Cluster-Aligned Prior (CAP) regularisation on matched positives, allowing prior information to stabilise metric size estimates without pulling all queries toward a single prototype.

  \item We evaluate MonoPRIO through official KITTI test results, matched five-run validation statistics, simple-prior, routing, and prior-strength controls, efficiency comparison, ambiguity-regime diagnostics, low-data experiments, quantitative out-of-support failure analysis, and additional dataset validation on the nuScenes frontal protocol.
\end{itemize}

\section{Related Work}

\subsection{Monocular 3D Detection}

Monocular 3D object detection must infer metric 3D structure from a single image, making it more under-constrained than stereo, LiDAR, or multi-view camera settings. Early monocular approaches therefore relied heavily on geometric structure and box parameterisation. Deep3DBox used geometric consistency between projected 3D boxes and 2D detections to recover object pose and dimensions \cite{mousavian20173d}, while earlier monocular detection formulations explored 3D proposal generation and geometric box lifting for autonomous driving scenes \cite{Chen_2016_CVPR}. Orthographic Feature Transform introduced an image-to-orthographic feature representation for monocular 3D reasoning \cite{roddick2018orthographic}. Later CNN-based detectors such as M3D-RPN and SMOKE moved toward practical anchor-, region-, and keypoint-based monocular pipelines \cite{Brazil_2019_ICCV,liu2020smoke}. These methods established strong monocular baselines, but their geometric cues remain sensitive to the inherent scale-depth ambiguity of single-view projection.

A second line of work improves monocular 3D detection by strengthening depth and geometric reasoning. D$^4$LCN uses depth-guided convolution to adapt image features to estimated depth structure \cite{ding2020learning}, while CaDDN models categorical depth distributions to support monocular 3D detection \cite{reading2021categorical}. GUPNet explicitly models uncertainty when projecting 2D observations into 3D, showing that confidence estimation is important for monocular depth recovery \cite{lu2021gupnet}. MonoDDE addresses the fact that a single image can support multiple plausible depths by modelling depth diversity during monocular 3D detection \cite{li2022monodde}. DEVIANT further studies how depth-related predictions should change consistently with image transformations, improving robustness under monocular geometric ambiguity \cite{kumar2022deviant}. OPA-3D combines dense depth and depth-to-box residual prediction with occlusion-aware geometric box recovery, using a fixed mean-size prior when pixel-wise geometric evidence is unreliable \cite{su2023opa3d}. These methods demonstrate the value of depth-aware geometric cues and structured priors, but their main emphasis is localisation, depth reliability, or geometric box recovery rather than query-adaptive conditioning of the metric size branch.

Recent monocular detectors further improve representation learning and decoder design. MonoCon adds auxiliary monocular contexts to support 3D attribute prediction \cite{liu2022monocon}. MonoUNI studies unified multi-class monocular detection, where a single detector must handle multiple object categories with different size and frequency characteristics \cite{jia2023monouni}. Transformer-based methods such as MonoDETR and MonoDGP improve query-based monocular reasoning through depth-guided decoding, decoupled query design, and geometry-aware modelling \cite{zhang2023monodetr,pu2025monodgp}. FD3D uses foreground depth information to strengthen monocular features for 3D detection \cite{wu2024fd3d}. MonoCD improves depth reasoning through complementary depth design \cite{yan2024monocd}. MonoMAE uses depth-aware masked autoencoding to strengthen monocular representations under occlusion and incomplete evidence \cite{jiang2024monomae}. Efficiency-aware designs such as MonoPoly and MonoA$^2$ explore scalable monocular detection and adaptive depth or multi-task heads \cite{guan2022monopoly,dong2026monoa2}. These works improve monocular features, depth cues, object queries, or efficiency, while MonoPRIO focuses on the narrower mechanism-level issue of stabilising metric size estimates through branch-local prior conditioning.

\subsection{Contextual and Adaptive Reasoning}

Recent methods increasingly improve monocular 3D detection by enriching query features or modelling dependencies between 3D attributes. MonoCLUE augments decoder queries with local clustering and scene-level memory, improving robustness under partial visibility and ambiguous local evidence \cite{yang2025monoclue}. MonoCoP combines chain-of-prediction feature propagation with an uncertainty-guided selector between sequential and parallel attribute prediction \cite{zhang2025monocop}. MonoDGP also shows that geometry-error modelling, segmentation-guided features, and decoupled query initialisation can improve monocular reasoning in a transformer detector \cite{pu2025monodgp}. These methods improve context modelling, attribute coupling, or depth and geometric reasoning. MonoPRIO is complementary: it does not introduce a broader scene-memory module or attribute chain, but instead applies query-adaptive prior conditioning directly to the size pathway.

\subsection{Camera-Only BEV and Height-Based Representations}

Camera-based 3D perception has also advanced through image-to-BEV and height-aware representation learning. DETR3D detects 3D objects from multi-view images using 3D-to-2D queries, linking object queries to image features through camera geometry \cite{wang2022detr3d}. BEVFormer learns a unified bird's-eye-view representation from multi-camera images using spatial cross-attention and temporal self-attention over BEV queries \cite{li2022bevformerlearningbirdseyeviewrepresentation}. BEVDepth improves camera-based BEV detection by strengthening intermediate depth estimation with explicit depth supervision and camera-aware depth modelling \cite{li2023bevdepth}. BEVHeight and BEVHeight++ address visual-centric and roadside 3D detection by replacing or supplementing depth-centric projection with height-to-ground modelling, improving the robustness of 2D-to-BEV projection under roadside camera geometry \cite{yang2023bevheight, yang2025bevheightpp}.

These BEV and height-based approaches address camera-only 3D perception by improving the representation used to project image features into 3D or BEV space. MonoPRIO addresses a different part of the monocular detection pipeline. It does not construct a BEV feature map, replace depth reasoning, or require a multi-camera setting. Instead, it conditions the object-level size branch of a monocular detector using query-routed class-aware size-prior statistics. The two directions are therefore complementary: BEV/height methods improve geometric representation learning, while MonoPRIO targets metric size stabilisation within an object-query detector.

\subsection{Priors, Uncertainty, and Size Stabilisation}

Priors and uncertainty estimates are widely used to reduce monocular ambiguity, but they are usually tied to depth, object placement, or general geometric consistency. As discussed above, existing methods use uncertainty-aware projection, depth-diversity modelling, placement priors, or geometry-aware query decoding to improve localisation and depth reliability. These approaches show that structured geometric information is valuable for monocular 3D detection, but they do not explicitly treat metric size as the direct target of adaptive prior conditioning.

MonoPRIO differs by constructing an offline class-aware size prototype bank and routing each decoder query to a soft mixture of prototype statistics. The prior is injected in log-size space and attenuated by routed uncertainty, while CAP regularisation encourages matched positives to remain near plausible prototype manifolds during training. This makes the contribution branch-local and size-specific: MonoPRIO does not aim to replace depth estimation or global scene modelling, but to stabilise metric size prediction when single-view evidence admits multiple plausible size-depth interpretations.

\section{Method}

\subsection{MonoPRIO Overview}
MonoPRIO targets metric size instability under partial evidence in unified multi-class detection through offline class-aware prior-bank construction, query-level mixture routing, uncertainty-aware log-space size conditioning, and CAP regularisation on matched positives.

\begin{figure}[t]
  \centering
  
    \includegraphics[width=\textwidth]{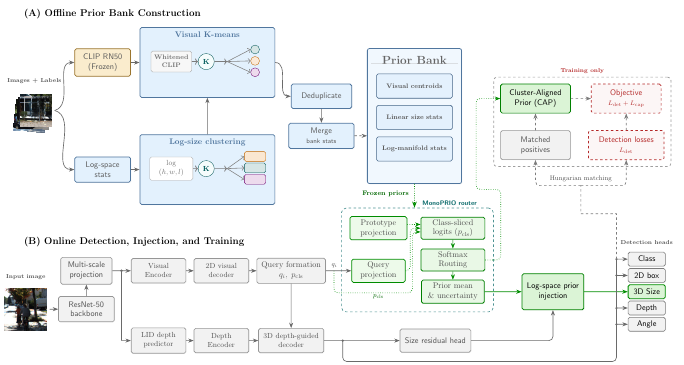}

  \caption{MonoPRIO pipeline. The class-aware prior bank is constructed offline from labelled training instances and frozen before detector training. During detection, decoder queries route to stored prototype statistics for uncertainty-aware log-space size conditioning. CAP regularisation is applied only to matched positives during training.}

  \label{fig:monoprio_system}
\end{figure}

Figure~\ref{fig:monoprio_system} summarises the full pipeline. At inference, each decoder query is routed to class-consistent prototype statistics, producing a query-specific prior mean and uncertainty for log-space size conditioning.

\subsection{Base Detector}
MonoPRIO builds on a MonoDGP backbone \cite{pu2025monodgp} and modifies only the size pathway while keeping the matching strategy and remaining prediction heads unchanged. This restriction is deliberate: metric size is the attribute most directly entangled with scale-depth ambiguity, so targeting this branch isolates the effect of prior conditioning and keeps attribution clear in ablations.

Let the final 2D decoder query for instance $i$ be $\mathbf{q}_i\in\mathbb{R}^D$, with normalised class-gating vector $\mathbf{p}_i$. MonoPRIO uses $(\mathbf{q}_i,\mathbf{p}_i)$ to route a query-specific size prior, then fuses it into the size residual head in log-space. We attach routing at this final 2D query because it is the most object-centric, visually grounded representation shared by the downstream 3D heads, enabling branch-local size conditioning without modifying the remaining detection pathway.

Although MonoPRIO modifies only the size pathway, its effect is not confined to the standalone dimension estimate. In the MonoDGP base detector, geometric depth is computed from the predicted metric height, projected 2D box height, and focal length:
\begin{equation}
  z_i^{\mathrm{geo}} =
  \frac{\hat{h}_i}{h_i^{\mathrm{2D}}} f_i,
\end{equation}
where $\hat{h}_i$ is the predicted metric height, $h_i^{\mathrm{2D}}$ is the projected 2D box height, and $f_i$ is the camera focal length. Prior-conditioned size predictions therefore affect the 3D box extent directly and can also affect depth and 3D localisation indirectly. Consequently, small changes in average dimension error can produce more visible changes in AP when they improve box alignment for detections near the evaluation IoU threshold.

\subsection{Offline Class-Aware Prior Bank}
We build class-specific prior banks offline from the detector training split of the target dataset. Each labelled instance is filtered by class-aware visibility and truncation thresholds. The corresponding crop is encoded by a frozen CLIP RN50 image encoder \cite{radford2021clip}. The CLIP encoder is executed only during offline bank construction. During detector training and inference, the encoder is not executed; routing uses only the stored CLIP-derived prototype centroids and associated bank statistics.

This offline design separates geometric and visual sources of variation before detector training. Metric dimensions define the physical modes that the size branch should respect, while visual appearance helps distinguish subtypes within those modes, such as viewpoint and residual intra-class shape variation. The bank is therefore constructed to anchor prototypes in metric size space before appearance is used for refinement. This reduces the risk of grouping instances primarily by appearance, since visually similar instances can still differ in physical dimensions. Keeping the CLIP encoder outside detector training and inference prevents it from becoming an additional train-time or test-time feature extractor, so the online detector remains lightweight and the effects of routing and CAP regularisation can be isolated in ablations.

For instance size $d_i=(h_i,w_i,l_i)$, we operate in log-space:
\begin{equation}
  \tilde d_i = \log(d_i + \epsilon).
\end{equation}

We first cluster $\tilde d_i$ in standardised log-size space to form coarse geometry groups of similar physical dimensions. Within each geometry group, we then cluster whitened CLIP features to form appearance-consistent subtypes. This geometry-first then appearance-second strategy separates physically distinct size modes before finer visual subtype partitioning, reducing cross-mode mixing relative to a single-stage clustering pass. Near-duplicate visual prototypes are merged using a class-specific cosine threshold. Their member sets are pooled and prototype statistics are recomputed from the union. Groups below a minimum support threshold are then assigned to the nearest retained prototype by visual-centroid similarity. For this low-support merge, the linear mean, linear standard deviation, and log-space mean use count-weighted averaging, a support-weighted approximation rather than pooled recomputation. The retained visual centroid and log-space basis are preserved, while inverse log standard deviations use the element-wise minimum, preventing increased precision along the retained axes. Each final prototype stores linear moments $(\mu_k,\sigma_k)$ and log-space manifold parameters $(\mu_k^{\log}, \Sigma_k^{\log})$, where

\begin{equation}
  \Sigma_k^{\log} = V_k^{\log} \, \mathrm{diag}(\eta_k) \, V_k^{\log\top}.
\end{equation}

Here, $V_k^{\log}$ is the stored orthonormal log-space basis and $\eta_k$ denotes the effective axis variances implied by the stored inverse log-standard-deviation scales. Before low-support merging, the basis is obtained from the empirical log-space covariance eigendecomposition and the axis scales are regularised by the stated eigenvalue floor and maximum inverse-standard-deviation constraint. After merging, the retained basis and conservatively updated scales define the effective manifold. The construction recipe is deterministic and uses only the detector training split for the corresponding experimental setting. No validation or test labels are used during prior-bank construction. For each retained instance, the 2D ground-truth crop is extracted from the image, padded to a square canvas without additional context expansion, and processed by the frozen CLIP RN50 image encoder using the standard CLIP preprocessing. CLIP features are $\ell_2$-normalised. PCA whitening is used only for the visual K-means stage, while stored visual centroids are recomputed in the original normalised CLIP feature space. Linear statistics $(\mu_k,\sigma_k)$ are used for routed prior formation, while log-space manifold statistics $(V_k^{\log},\eta_k,\mu_k^{\log})$ define the CAP geometry. The bank is rebuilt once for each training split or dataset setting, frozen before detector training, and reused across repeated runs.

\begin{table}[htbp]
  \centering
  \begin{singlespace}
  \caption{Prior-bank construction details for KITTI. Initial geometry groups are formed by K-means in standardised log-size space. Visual groups are formed by K-means on whitened CLIP features within each geometry group. Final prototype counts are after near-duplicate merging and low-support merging.}
  \label{tab:prior_bank_details}
  \small
  \renewcommand{\arraystretch}{1.18}
  \begin{tabular*}{0.88\linewidth}{@{\extracolsep{\fill}}lccc@{}}
  \toprule
  Setting & Car & Pedestrian & Cyclist \\
  \midrule
  Final prototypes & 10 & 13 & 12 \\
  \addlinespace[1pt]
  Initial geometry groups $K_{\mathrm{geo}}$ & 5 & 4 & 4 \\
  Visual groups per geometry group $K_{\mathrm{vis}}$ & 4 & 2 & 2 \\
  \addlinespace[1pt]
  Minimum 2D box height & 30 px & 15 px & 15 px \\
  Maximum occlusion level & 0 & 2 & 2 \\
  Maximum truncation & 0.0 & 0.3 & 0.3 \\
  \addlinespace[1pt]
  Minimum cluster support & 10 & 30 & 12 \\
  Deduplication cosine threshold & 0.999 & 0.9995 & 0.9995 \\
  \addlinespace[1pt]
  Minimum linear size std. & 0.03 & 0.05 & 0.05 \\
  Minimum log covariance eigenvalue & $10^{-3}$ & 0.002 & 0.002 \\
  Maximum inverse log std. & 10.0 & 6.0 & 8.0 \\
  \bottomrule
  \end{tabular*}
  \end{singlespace}
\end{table}

Prior banks are generated with seed 42, and K-means uses 10 initialisations. The unified bank follows the class-slice order Pedestrian, Car, Cyclist, matching the detector class-index convention, and contains 35 prototypes in total. The prior-bank construction hyperparameters are fixed across all repeated runs within a dataset setting and are not tuned per validation seed. Online prior-strength sensitivity is evaluated using $0.5\times$, $1\times$, and $2\times$ the default $\lambda_0$, while keeping the prior bank, routing mechanism, uncertainty attenuation, class scales, and CAP settings fixed.

\subsection{Query Routing and Uncertainty-Aware Mixture Prior}
Given query $\mathbf{q}_i$ and stored prototype visual centroid $\mathbf{v}_k$, we project and normalise both:

\begin{equation}
  \mathbf{q}_i' = \frac{W_q \mathbf{q}_i}{\|W_q \mathbf{q}_i\|}, \quad
  \mathbf{v}_k' = \frac{W_k \mathbf{v}_k}{\|W_k \mathbf{v}_k\|}.
\end{equation}

Routing logits are
  \begin{equation}
  \ell_{ik} = \alpha \mathbf{q}_i^{\prime\top}\mathbf{v}'_k,
  \end{equation}
where $\alpha=1/\sqrt{256}$ is a fixed logit scale. Let $z_{ic}$ denote the final 2D-decoder classification logit for query $i$ and class $c$. Because the base detector uses sigmoid focal classification, the class-gating weights are formed as

\begin{equation}
  p_{ic} =
  \frac{\operatorname{sigmoid}(z_{ic})}
  {\sum_{c'=1}^{C}\operatorname{sigmoid}(z_{ic'})}.
\end{equation}

The denominator is clamped by $\epsilon$ in implementation for numerical stability. The resulting vector $\mathbf{p}_i$ is detached before routing, so the prior pathway consumes semantic confidence without propagating prior-loss gradients into the classification branch. These weights gate routing over class slices $\mathcal{S}_c$:

\begin{equation}
    \tilde{a}_{ik} =
    \begin{cases}
      \operatorname{softmax}(\boldsymbol{\ell}_{i,\mathcal{S}_c})_k\,p_{ic},
      & k \in \mathcal{S}_c,\\
      0, & \text{otherwise},
    \end{cases}
    \qquad
    a_{ik} =
    \frac{\tilde{a}_{ik}}
    {\sum_j \tilde{a}_{ij}}.
\end{equation}

Class gating first restricts prototype competition within each class slice and then allocates the resulting routing mass across slices according to the normalised class-gating weights. Routing therefore remains probabilistic when class confidence is diffuse while concentrating on one class slice as classification confidence sharpens.

Because $\mathbf{q}'_i$ and $\mathbf{v}'_k$ are $\ell_2$-normalised, their dot product is bounded, and the fixed logit scale keeps each within-class routing distribution diffuse rather than approaching a hard prototype assignment. We therefore interpret $a_{ik}$ as soft mixture weights used to form routed prior statistics, rather than as calibrated probabilities over discrete prototype identities. The learned projections can alter the relative ordering of prototype weights; during the later differentiable CAP stage, CAP gradients can also influence this ordering. We therefore analyse soft-distribution sharpness separately from the frequency with which a prototype attains the largest bank-wide routing weight. Diffuse routing does not imply a class-constant prior. Query-dependent changes in $a_i$ yield different $\hat{\mu}_i$ and $\hat{\sigma}_i$ whenever prototype statistics differ. We therefore assess routing specificity through the resulting mixture statistics rather than requiring sharply peaked assignments.

The routed mixture prior is

\begin{equation}
  \begin{aligned}
    \hat\mu_i &= \sum_k a_{ik}\mu_k, \\
    m_{2,i} &= \sum_k a_{ik}(\sigma_k^2+\mu_k^2), \\
    \hat\sigma_i &= \sqrt{\max(m_{2,i}-\hat\mu_i^2,\epsilon)}.
  \end{aligned}
\end{equation}

The mixture statistics $\hat\mu_i$ and $\hat\sigma_i$ are computed element-wise over $(h,w,l)$. The expression for $\hat\sigma_i$ is the finite-mixture law of total variance, combining within-prototype spread $\sigma_k^2$ with disagreement between prototype means. $\hat\sigma_i$ is therefore the query-level routed uncertainty.

This routing formulation couples semantic plausibility with instance adaptation. Class gating allocates routing mass across class slices according to semantic confidence, soft within-slice mixtures avoid brittle hard assignment, and $\hat{\sigma}_i$ provides a query-level uncertainty signal for downstream prior-strength control. In implementation terms, $W_q$ and $W_k$ are learned jointly with the detector, whereas bank statistics $(\mu_k,\sigma_k,\mu_k^{\log},V_k^{\log},\eta_k)$ are frozen after offline construction, prior to detector training.

\subsection{Log-Space Prior-Conditioned Size Prediction}
The size head predicts a residual vector $r_i$ in log-space. We inject the routed prior mean as

\begin{equation}
  \hat s_i = \exp\left(r_i + \lambda_i \odot \log(\hat\mu_i+\epsilon)\right),
  \quad
  \lambda_i = \lambda_0\,c_i\,g_i,
\end{equation}

where $\lambda_0$ is a base prior-strength hyperparameter, $c_i=\sum_c p_{ic}\beta_c^{\mathrm{cls}}$ is class-conditioned scaling, and $g_i=(1+\hat\sigma_i/\sigma_s)^{-1}$ attenuates prior influence under high uncertainty, with $\sigma_s$ a fixed uncertainty-scale hyperparameter. Products and divisions are applied element-wise over $(h,w,l)$. This parameterisation preserves positivity while retaining instance-specific residual freedom around routed priors, instead of collapsing predictions toward a rigid class mean. Operating in log-space also makes conditioning additive in a space that better matches multiplicative size variation than direct linear correction.

\subsection{Cluster-Aligned Prior (CAP) Regularisation}\label{subsec:cap_regularisation}
CAP shapes the optimisation landscape during training, while routed injection modifies the predicted size estimate. CAP is applied only on matched positives, so it regularises plausible assignments rather than all queries indiscriminately. This restriction keeps CAP aligned with the detector's existing supervision pathway. The prior bank is intended to shape metric size estimates for detections that have already been assigned to ground-truth objects, rather than to impose object-like dimensions on unmatched background queries. Applying the loss only after detector matching therefore avoids coupling prior regularisation to low-confidence or background predictions.

CAP is also a soft prototype-manifold regulariser rather than a hard nearest-prototype constraint. A matched instance can remain compatible with multiple plausible size modes, so CAP weights prototype distances by the routed assignment distribution instead of forcing a single discrete choice. This preserves instance-specific residual freedom in the size head while discouraging predictions from drifting outside plausible class-conditioned size manifolds.

Let $x_i=\log(\hat s_i)$ for matched prediction $i$. The prototype-wise whitened distance is

\begin{equation}
  \mathrm{md}_{ik}^2=
  \left\|\left(V_k^{\log\top}(x_i-\mu_k^{\log})\right)\odot\eta_k^{-1/2}\right\|_2^2.
\end{equation}

Projection by $V_k^{\log\top}$ followed by the stored inverse axis scaling $\eta_k^{-1/2}$ therefore evaluates the Mahalanobis distance under the covariance implied by the stored manifold parameters.

The CAP loss is

\begin{equation}
  \mathcal{L}_{\mathrm{cap}}=
  \frac{1}{N_{+}}\sum_{i=1}^{N_{+}} w^{\mathrm{cap}}_{y_i}\sum_k a_{ik}\,\mathrm{md}_{ik}^2,
\end{equation}

where $w^{\mathrm{cap}}_{y_i}$ is a class-dependent CAP weight for ground-truth class $y_i$, and $N_{+}$ is the number of matched positives in the mini-batch. Weighting by $a_{ik}$ keeps CAP soft, avoiding brittle attraction to a single prototype when multiple plausible size modes remain consistent with the query. For the CAP weighting path only, routing weights use staged gradient flow: detached in early epochs, linearly blended detached/non-detached in a transition phase, and fully differentiable in later epochs.

The full training objective is then

\begin{equation}
  \mathcal{L}=\mathcal{L}_{\mathrm{det}}+\lambda_{\mathrm{cap}}\,\rho(e)\,\mathcal{L}_{\mathrm{cap}},
\end{equation}

where $\rho(e)$ keeps CAP at full strength in early training and then linearly reduces CAP weight in later epochs. All prior-bank construction is offline; at inference, MonoPRIO uses only stored bank statistics and lightweight routing in the size branch.

\section{Experimental Setup}

\subsection{Dataset and Protocols}
We primarily evaluate on KITTI~\cite{geiger2012kitti}. Following common practice~\cite{NIPS2015_6da37dd3}, we use the 3,712/3,769 train/val split for local analysis. Benchmark results are reported on the official KITTI test server. Car is evaluated at IoU 0.7, while Pedestrian/Cyclist are evaluated at IoU 0.5. Local validation results are reported as medians over five runs. We additionally report interquartile ranges (Q1--Q3) for the primary validation metrics, key size-path ablations, and low-data UniMod results.

We additionally evaluate on the nuScenes frontal protocol used by DEVIANT and MonoCoP~\cite{kumar2022deviant,zhang2025monocop}. Following this protocol, we use the official nuScenes v1.0-trainval train and validation splits and the front-facing CAM\_FRONT keyframes converted into KITTI style. The converted data contain 28,130 training images and 6,019 validation images. Because the conversion assigns zero truncation and occlusion to the annotations, Easy and Moderate are determined by 2D bounding-box heights of at least 40 and 25 pixels, respectively. We report Car AP$_{3D|R40}$ and AP$_{BEV|R40}$ at IoU thresholds of 0.7 and 0.5. A Car-only prior bank is rebuilt from the converted nuScenes training split and contains 11 final prototypes. No validation annotations are used during bank construction. This experiment therefore evaluates transfer of the prior-bank construction and conditioning mechanism rather than direct transfer of a KITTI prior bank.

\subsection{Baselines and Metrics}\label{subsec:metrics}
We compare against representative monocular methods including MonoDGP \cite{pu2025monodgp}, MonoCLUE \cite{yang2025monoclue}, MonoCoP \cite{zhang2025monocop}, and the additional monocular baselines listed in Tables~\ref{tab:kitti_test_val_combined} and~\ref{tab:car_test_val_monoprio}. Primary metrics are AP$_{3D|R40}$ and AP$_{BEV|R40}$. We report unified moderate-score average (UniMod), size MAE on matched true positives, and Outlier ratio@20\% for mechanism analysis.

\begin{equation}
\begin{aligned}
  \mathrm{UniMod}=
  \frac{1}{3}\big(&\mathrm{AP}^{\mathrm{Car}}_{3D,\mathrm{Mod}@0.7}
  +\mathrm{AP}^{\mathrm{Ped}}_{3D,\mathrm{Mod}@0.5} \\
  &+\mathrm{AP}^{\mathrm{Cyc}}_{3D,\mathrm{Mod}@0.5}\big).
\end{aligned}
\end{equation}

For mechanism analysis, size MAE is the mean absolute error over $(h,w,l)$ on matched true positives. Rel MAE is the corresponding mean relative error, and Outlier ratio@20\% is the fraction of matched true positives whose mean relative error exceeds 20\%. For the direct three-method comparison in Table~\ref{tab:monoprio_mechanism_ablation}, the size diagnostics are computed on the per-run intersection of ground-truth instances matched as true positives by all three methods, identified by image, class, and ground-truth annotation index.

These metrics are complementary. AP measures final detection quality after confidence ranking, localisation, depth, and size estimation are combined, whereas the matched true-positive size metrics isolate the behaviour of the size branch more directly. This distinction is important because MonoPRIO is designed to stabilise metric size rather than to replace the full detection pipeline. UniMod provides a compact summary of the unified multi-class setting, while the class-wise AP, size MAE, relative error, and outlier analyses show whether gains arise from broad detection improvements or from reduced size instability in ambiguity-prone cases.

\subsection{Implementation Details}

For repeated local analysis, each method is evaluated under the same fixed training and evaluation schedule within the corresponding setting. Reported validation results are medians over five matched runs, and component comparisons use seed-wise pairing. Low-data experiments use the final logged evaluation for the matched runs. The official KITTI test submission was trained on the full KITTI training set and uses predictions from epoch 250 without split-specific retuning. Offline prior banks are fixed per training split and reused across repeated runs. For simple-prior controls, the class-mean variant uses one prototype per class, the size-only baseline removes CLIP visual features and uses uniform within-class routing, hard routing replaces the soft mixture with argmax prototype assignment, the no-uncertainty-gate variant disables uncertainty attenuation, and the CAP-only variant disables log-space size injection while retaining CAP regularisation. The VS/uniform control retains the full visual-size bank, class gating, uncertainty attenuation, size injection, and CAP settings, but replaces learned within-class routing weights with uniform weights. The prior-strength sweep varies only $\lambda_0$ over $0.5\times$, $1\times$, and $2\times$ its default value. The routed injection and Injection + CAP variants use the same prior bank, so their difference isolates the contribution of CAP; comparisons with MonoDGP retain the same base detector and training schedule. Runtime, FLOPs, and parameter counts are measured under identical software, hardware, input-resolution, and batch-size settings on a single RTX 5090 GPU using batch size 1 and $384{\times}1280$ input resolution.

For the default full-data KITTI configuration, we set $\lambda_0=0.058$, $\sigma_s=0.0077$, and $\beta^{\mathrm{cls}}=(0.142,0.154,0.120)$ for Pedestrian, Car, and Cyclist, respectively. We set $\lambda_{\mathrm{cap}}=0.0052$ and the corresponding class-dependent CAP weights to $(0.020,0.024,0.014)$ in the same class order. Within the CAP loss, routing weights are detached during epochs 0--19, linearly blended between detached and differentiable weights during epochs 20--39, and fully differentiable from epoch 40 onward. The CAP schedule keeps $\rho(e)=1$ through epoch 92 and then decays it linearly to zero at epoch 199, remaining zero thereafter.

\section{Experimental Results and Analysis}

\subsection{KITTI Main Results: Official Test and Unified Validation}

Table~\ref{tab:kitti_test_val_combined} combines published official-test results with five-run local validation medians obtained under our matched unified-training protocol. Within each split block, methods are ordered by Car AP$_{3D|R40}$ Moderate where reported. Among the compared official test methods reporting complete Car, Pedestrian, and Cyclist results, the unified MonoPRIO model achieves the highest Moderate AP in all three categories. MonoPRIO reaches 18.93, 10.74, and 4.32 Moderate AP$_{3D|R40}$ for Car, Pedestrian, and Cyclist, respectively, compared with 18.70, 10.45, and 3.20 for MonoCLUE. Local five-run validation medians also favour MonoPRIO under the matched unified protocol. These results are consistent with the main empirical pattern of the paper: query-adaptive prior conditioning is especially beneficial in the unified multi-class regime, where class imbalance and projection ambiguity are jointly present.

\begin{table}[htbp]
  \centering
  \caption{KITTI official-test and unified-validation results. Car uses IoU=0.7; Pedestrian/Cyclist use IoU=0.5. Test values are published official benchmark results, while validation values are five-run medians under our unified protocol. Claims about complete multi-class performance are restricted to methods reporting all three categories. Best and second-best values within each split are shown in \textbf{bold} and \underline{underlined}; $\dagger$ denotes additional depth supervision.}
  \label{tab:kitti_test_val_combined}
  \scriptsize
  \setlength{\tabcolsep}{2.2pt}
  \renewcommand{\arraystretch}{1.25}
  \resizebox{\textwidth}{!}{%
  \begin{tabular}{c|l|ccc|ccc|ccc|ccc}
    \toprule
    Split & Method &
    \multicolumn{3}{c|}{Car AP$_{BEV|R40}$} &
    \multicolumn{3}{c|}{Car AP$_{3D|R40}$} &
    \multicolumn{3}{c|}{Ped. AP$_{3D|R40}$} &
    \multicolumn{3}{c}{Cyc. AP$_{3D|R40}$} \\
    \cmidrule(lr){3-5}\cmidrule(lr){6-8}\cmidrule(lr){9-11}\cmidrule(lr){12-14}
    & & Easy & Mod. & Hard & Easy & Mod. & Hard & Easy & Mod. & Hard & Easy & Mod. & Hard \\
    \midrule
    \multirow{12}{*}{Test}
    & GUPNet~\cite{lu2021gupnet} & -- & -- & -- & -- & -- & -- & 14.72 & 9.53 & 7.87 & 4.18 & 2.65 & 2.09 \\
    & DEVIANT~\cite{kumar2022deviant} & 29.65 & 20.44 & 17.43 & 21.88 & 14.46 & 11.89 & 13.43 & 8.65 & 7.69 & 5.05 & 3.13 & 2.59 \\
    & MonoCon~\cite{liu2022monocon} & 31.12 & 22.10 & 19.00 & 22.50 & 16.46 & 13.95 & 13.10 & 8.41 & 6.94 & 2.80 & 1.92 & 1.55 \\
    & MonoPoly-L~\cite{guan2022monopoly} & -- & -- & -- & -- & -- & -- & 11.05 & 7.74 & 6.94 & 5.73 & 3.40 & 2.79 \\
    & MonoUNI~\cite{jia2023monouni} & 33.28 & 23.05 & 19.39 & 24.75 & 16.73 & 13.49 & 15.78 & 10.34 & 8.74 & 7.34 & 4.28 & \underline{3.78} \\
  & OPA-3D$^{\dagger}$~\cite{su2023opa3d} & 33.54 & 22.53 & 19.22 & 24.60 & 17.05 & 14.25 & -- & -- & -- & -- & -- & -- \\
    & MonoDDE~\cite{li2022monodde} & 33.58 & 23.46 & 20.37 & 24.93 & 17.14 & 15.10 & 11.13 & 7.32 & 6.67 & 5.94 & 3.78 & 3.33 \\
    & MonoDGP~\cite{pu2025monodgp} & 35.19 & 24.50 & 21.29 & 26.05 & 18.55 & 15.95 & 15.04 & 9.89 & 8.38 & 5.28 & 2.82 & 2.65 \\
    & MonoCoP~\cite{zhang2025monocop} & -- & -- & -- & -- & -- & -- & 15.61 & 10.33 & 8.53 & \textbf{8.89} & \textbf{5.08} & \textbf{5.25} \\
    & MonoA$^2$~\cite{dong2026monoa2} & -- & -- & -- & -- & -- & -- & 12.95 & 8.51 & 7.56 & 4.39 & 2.28 & 2.31 \\
    & MonoCLUE~\cite{yang2025monoclue} & \textbf{35.83} & \underline{24.59} & \underline{21.36} & \underline{26.59} & \underline{18.70} & \underline{15.99} & \underline{16.18} & \underline{10.45} & \underline{8.75} & 5.93 & 3.20 & 2.94 \\
    & \textbf{MonoPRIO (Ours)} & \underline{35.50} & \textbf{24.66} & \textbf{21.54} & \textbf{26.83} & \textbf{18.93} & \textbf{16.25} & \textbf{16.31} & \textbf{10.74} & \textbf{9.08} & \underline{7.72} & \underline{4.32} & 3.61 \\
    \midrule
    \multirow{3}{*}{Val}
    & MonoDGP~\cite{pu2025monodgp} & 37.451 & 26.938 & 24.416 & \underline{29.536} & 21.136 & \underline{18.802} & \underline{12.311} & \underline{9.270} & \underline{7.268} & 9.813 & 4.921 & 4.422 \\
    & MonoCLUE~\cite{yang2025monoclue} & \underline{38.249} & \underline{28.119} & \underline{24.656} & 29.344 & \underline{21.416} & 18.495 & 11.658 & 8.579 & 7.043 & \underline{11.180} & \underline{5.741} & \underline{5.259} \\
    & \textbf{MonoPRIO (Ours)} & \textbf{38.390} & \textbf{28.538} & \textbf{24.723} & \textbf{30.413} & \textbf{21.856} & \textbf{18.965} & \textbf{12.408} & \textbf{9.361} & \textbf{7.332} & \textbf{12.402} & \textbf{5.988} & \textbf{5.899} \\
    \bottomrule
  \end{tabular}}
\end{table}

\begin{table}[htbp]
  \centering
  \begin{singlespace}
  \caption{Run-to-run dispersion for primary KITTI validation metrics. Values are AP$_{3D|R40}$ Moderate median [Q1, Q3] over five runs. UniMod is computed per run as the mean of Car, Pedestrian, and Cyclist Moderate AP$_{3D|R40}$.}
  \label{tab:kitti_val_dispersion}
  \small
  \setlength{\tabcolsep}{4.0pt}
  \renewcommand{\arraystretch}{1.15}
  \begin{tabular*}{\textwidth}{@{\extracolsep{\fill}}lcccc@{}}
  \toprule
  Method & Car & Ped. & Cyc. & UniMod \\
  \midrule
  MonoDGP
  & \makecell[c]{21.136\\{\scriptsize [21.074, 22.615]}}
  & \makecell[c]{9.270\\{\scriptsize [8.393, 9.885]}}
  & \makecell[c]{4.921\\{\scriptsize [4.805, 5.158]}}
  & \makecell[c]{12.105\\{\scriptsize [11.951, 12.415]}} \\
  MonoCLUE
  & \makecell[c]{21.416\\{\scriptsize [21.363, 21.823]}}
  & \makecell[c]{8.579\\{\scriptsize [7.080, 9.275]}}
  & \makecell[c]{5.741\\{\scriptsize [5.194, 6.537]}}
  & \makecell[c]{11.412\\{\scriptsize [11.249, 12.576]}} \\
  \textbf{MonoPRIO}
  & \makecell[c]{\textbf{21.856}\\{\scriptsize [21.700, 21.927]}}
  & \makecell[c]{\textbf{9.361}\\{\scriptsize [9.176, 9.652]}}
  & \makecell[c]{\textbf{5.988}\\{\scriptsize [5.893, 6.791]}}
  & \makecell[c]{\textbf{12.307}\\{\scriptsize [12.250, 12.608]}} \\
  \bottomrule
  \end{tabular*}
  \end{singlespace}
\end{table}

Table~\ref{tab:kitti_val_dispersion} reports dispersion for the primary validation metrics. MonoPRIO has the highest median AP across Car, Pedestrian, Cyclist, and UniMod, while the interquartile ranges show the accompanying run-to-run variability across five runs.

\subsection{Secondary Car-Only Evaluation}
Table~\ref{tab:car_test_val_monoprio} is reported as a secondary setting to characterise boundary conditions for the prior-conditioning mechanism. In car-only training, inter-class ambiguity is reduced, so class-aware routed priors are less central than in unified multi-class detection. MonoPRIO is stronger than MonoCLUE on the official KITTI car-only test benchmark across all reported Car AP$_{3D|R40}$ difficulty levels. This secondary setting indicates competitive car-only behaviour, while the clearest MonoPRIO gains occur in the unified multi-class regime.

\begin{table}[htbp]
\centering
\begin{singlespace}
  \caption{Car-only KITTI official-test results at IoU=0.7. Methods are grouped by their use of additional LiDAR or depth supervision. MonoPRIO uses frozen CLIP RN50 only for offline prior-bank construction. Best and second-best values among methods without additional 3D supervision are shown in \textbf{bold} and \underline{underlined}.}
  \label{tab:car_test_val_monoprio}
\scriptsize
\setlength{\tabcolsep}{2.4pt}
\renewcommand{\arraystretch}{1.02}
\setlength{\cmidrulewidth}{0.45pt}
\resizebox{0.95\textwidth}{!}{%
\begin{tabular}{llccc@{\hspace{10pt}}ccc}
    \toprule
    Method & Extra 3D sup. &
    \multicolumn{3}{c}{Car AP$_{BEV|R40}$} &
    \multicolumn{3}{c}{Car AP$_{3D|R40}$} \\
    \cmidrule(lr){3-5}\cmidrule(lr){6-8}
    & & Easy & Mod. & Hard & Easy & Mod. & Hard \\
    \midrule
    MonoDTR~\cite{huang2022monodtr} & LiDAR & 28.59 & 20.38 & 17.14 & 21.99 & 15.39 & 12.73 \\
    DID-M3D~\cite{peng2022didm3d} & LiDAR & 32.95 & 22.76 & 19.83 & 24.40 & 16.29 & 13.75 \\
    OccupancyM3D~\cite{peng2024occupancym3d} & LiDAR & 35.38 & 24.18 & 21.37 & 25.55 & 17.02 & 14.79 \\
    MonoTAKD~\cite{liu2025monotakd} & LiDAR & 38.75 & 27.76 & 24.14 & 27.91 & 19.43 & 16.51 \\
    MonoPGC~\cite{wu2023monopgc} & Depth & 32.50 & 23.14 & 20.30 & 24.68 & 17.17 & 14.14 \\
    OPA-3D~\cite{su2023opa3d} & Depth & 33.54 & 22.53 & 19.22 & 24.60 & 17.05 & 14.25 \\
    \midrule
    GUPNet~\cite{lu2021gupnet} & None & -- & -- & -- & 20.11 & 14.20 & 11.77 \\
    MonoFlex~\cite{zhang2021objects} & None & 28.23 & 19.75 & 16.89 & 19.94 & 13.89 & 12.07 \\
    MonoRCNN~\cite{shi2021geometry} & None & 25.48 & 18.11 & 14.10 & 18.36 & 12.65 & 10.03 \\
    DEVIANT~\cite{kumar2022deviant} & None & 29.65 & 20.44 & 17.43 & 21.88 & 14.46 & 11.89 \\
    MonoPoly-L~\cite{guan2022monopoly} & None & 29.57 & 22.23 & 19.75 & 22.94 & 16.13 & 13.01 \\
    MonoCon~\cite{liu2022monocon} & None & 31.12 & 22.10 & 19.00 & 22.50 & 16.46 & 13.95 \\
    MonoDETR~\cite{zhang2023monodetr} & None & 33.60 & 22.11 & 18.60 & 25.00 & 16.47 & 13.58 \\
    MonoUNI~\cite{jia2023monouni} & None & -- & -- & -- & 24.75 & 16.73 & 13.49 \\
    FD3D~\cite{wu2024fd3d} & None & 34.20 & 23.72 & 20.76 & 25.38 & 17.12 & 14.50 \\
    MonoCD~\cite{yan2024monocd} & None & 33.41 & 22.81 & 19.57 & 25.53 & 16.59 & 14.53 \\
    MonoMAE~\cite{jiang2024monomae} & None & 34.15 & 24.93 & 21.76 & 25.60 & 18.84 & \underline{16.78} \\
    MonoDGP~\cite{pu2025monodgp} & None & 35.24 & 25.23 & 22.02 & 26.35 & 18.72 & 15.97 \\
    MonoCoP~\cite{zhang2025monocop} & None & \textbf{36.77} & 25.57 & 22.62 & 27.54 & 19.11 & 16.33 \\
    MonoA$^2$~\cite{dong2026monoa2} & None & 31.71 & 23.14 & 20.45 & 23.24 & 17.55 & 15.26 \\
    MonoCLUE~\cite{yang2025monoclue} & None & \underline{36.15} & \textbf{26.15} & \underline{22.81} & \underline{27.94} & \underline{19.70} & 16.69 \\
    \textbf{MonoPRIO (Ours)} & None & 35.74 & \underline{26.05} & \textbf{22.90} & \textbf{28.12} & \textbf{19.85} & \textbf{16.93} \\
    \bottomrule
  \end{tabular}}
\end{singlespace}
\end{table}

\subsection{Additional Dataset Validation on nuScenes}

Table~\ref{tab:nuscenes_generalisation} reports additional validation under the nuScenes frontal protocol. For MonoPRIO, we report the best validation checkpoint from a 55-epoch training schedule, using the same checkpoint for all reported metrics. The training horizon and learning-rate milestones are rescaled for the larger nuScenes training split, while the prior-conditioning and CAP settings are adapted to the car-only configuration. The prior bank is rebuilt from the nuScenes training data and contains 11 Car prototypes.

\begin{table}[htbp]
    \centering
    \begin{singlespace}
    \caption{nuScenes frontal Car detection under the DEVIANT KITTI-style protocol. Values are AP$_{\mathrm{3D}|R40}$ and AP$_{\mathrm{BEV}|R40}$. For MonoPRIO, the best validation checkpoint from a 55-epoch training schedule is used for all reported metrics. Best and second-best values are shown in \textbf{bold} and \underline{underlined}.}
      \label{tab:nuscenes_generalisation}
    \begin{tabular}{llcccc}
        \toprule
        \multirow{2}{*}{Method} & \multirow{2}{*}{IoU$_{\mathrm{3D}}$} & \multicolumn{2}{c}{AP$_{\mathrm{3D}}$} & \multicolumn{2}{c}{AP$_{\mathrm{BEV}}$} \\
        \cmidrule(lr){3-4} \cmidrule(lr){5-6}
         & & Easy & Mod. & Easy & Mod. \\
        \midrule
        GUPNet \cite{lu2021gupnet} & \multirow{6}{*}{0.7} & 8.50 & 7.40 & 14.21 & 12.81 \\
        DEVIANT \cite{kumar2022deviant} & & 9.69 & 8.33 & 16.28 & 14.36 \\
        MonoDETR \cite{zhang2023monodetr} & & 9.53 & 8.19 & 16.39 & 14.41 \\
        MonoDGP \cite{pu2025monodgp} & & 10.04 & 8.78 & 16.55 & 14.53 \\
        \textbf{MonoPRIO (Ours)} & & \underline{10.39} & \underline{9.41} & \underline{17.54} & \underline{15.60} \\
        MonoCoP \cite{zhang2025monocop} & & \textbf{10.85} & \textbf{9.71} & \textbf{17.83} & \textbf{15.86} \\
        \midrule
        GUPNet \cite{lu2021gupnet} & \multirow{6}{*}{0.5} & 29.03 & 26.16 & 33.42 & 30.23 \\
        DEVIANT \cite{kumar2022deviant} & & 31.47 & 28.22 & 35.61 & 31.93 \\
        MonoDETR \cite{zhang2023monodetr} & & 31.81 & 28.35 & 35.70 & 31.96 \\
        MonoDGP \cite{pu2025monodgp} & & 29.56 & 26.17 & 32.67 & 29.44 \\
        \textbf{MonoPRIO (Ours)} & & \underline{33.17} & \underline{29.61} & \underline{36.90} & \underline{33.74} \\
        MonoCoP \cite{zhang2025monocop} & & \textbf{33.70} & \textbf{29.91} & \textbf{37.44} & \textbf{34.01} \\
        \bottomrule
    \end{tabular}
    \end{singlespace}
\end{table}

MonoPRIO improves over MonoDGP across all reported nuScenes metrics, with larger gains at IoU$_{\mathrm{3D}}=0.5$. These results show that the prior-bank construction and conditioning mechanism remains beneficial when rebuilt for nuScenes, rather than demonstrating direct transfer of a KITTI prior bank. MonoPRIO remains below MonoCoP, suggesting that branch-local size conditioning is complementary to broader attribute-coupling architectures.

\subsection{Efficiency and Accuracy-Cost Trade-off}

Table~\ref{tab:efficiency_unified} shows that MonoPRIO improves unified accuracy with near-baseline cost relative to MonoDGP: FLOPs are unchanged, parameters increase by only 0.33M, and latency rises by 1.24 ms under the matched protocol. Compared with MonoCLUE, MonoPRIO achieves higher UniMod with fewer parameters, lower FLOPs, and lower latency, consistent with the efficiency of branch-local size conditioning.

\begin{table}[htbp]
\centering
\begin{singlespace}
  \caption{Efficiency under unified accuracy reporting on the validation split. AP$_{3D|R40}$ Moderate values are medians over five runs. Runtime/FLOPs/parameters are measured on one RTX 5090 with batch size 1 and input resolution $384{\times}1280$. Best is \textbf{bold}, second-best is \underline{underlined}.}
  \label{tab:efficiency_unified}
  \normalsize
  \setlength{\tabcolsep}{5.0pt}
  \renewcommand{\arraystretch}{1.15}
  \begin{tabular*}{\textwidth}{@{\extracolsep{\fill}}lccccccc@{}}
    \toprule
    Method & Params$\downarrow$ & FLOPs$\downarrow$ & Latency$\downarrow$ & \multicolumn{4}{c}{AP$_{3D|R40}$ Mod.$\uparrow$} \\
    \cmidrule(lr){2-4}\cmidrule(lr){5-8}
    & (M) & (G) & (ms) & Car & Ped. & Cyc. & UniMod \\
    \midrule
    MonoDGP & \textbf{42.16} & \textbf{68.99} & \textbf{43.06} & 21.136 & \underline{9.270} & 4.921 & \underline{12.105} \\
    MonoCLUE & 44.17 & 72.71 & 68.57 & \underline{21.416} & 8.579 & \underline{5.741} & 11.412 \\
    \textbf{MonoPRIO} & \underline{42.49} & \textbf{68.99} & \underline{44.30} & \textbf{21.856} & \textbf{9.361} & \textbf{5.988} & \textbf{12.307} \\
    \bottomrule
  \end{tabular*}
\end{singlespace}
\end{table}

\subsection{Mechanism and Prior Ablation}

Table~\ref{tab:monoprio_mechanism_ablation} compares size-path behaviour on a common ground-truth set, retaining only instances matched as true positives by all three methods within each run (median 2,286 objects; range 2,252--2,458). Injection + CAP improves Cyclist MAE relative to both MonoDGP and Injection only, and improves unified MAE relative to Injection only, while MonoDGP retains slightly lower Car, Pedestrian, and unified MAE overall. Seed-wise pairing confirms that CAP is selective. Relative to Injection only, it improves Moderate AP$_{3D|R40}$ in 3/5 Car runs, 5/5 Pedestrian runs, and 3/5 Cyclist runs, but UniMod in 2/5 runs (median paired change $-0.041$). On the common-ground-truth set, Cyclist MAE improves in 3/5 runs and unified MAE in 2/5. We therefore treat CAP as complementary rather than uniformly beneficial. UniMod remains the detector-level five-run endpoint and is shown as supporting context.

Table~\ref{tab:simple_prior_ablation} adds matched bank-construction and routing controls. Under uniform routing, replacing the size-only bank with the visual-size bank yields modest improvements in unified MAE, outlier rate, and UniMod. Learned soft routing over the same visual-size bank increases UniMod from 11.861 to 12.307 and lowers unified MAE from 0.1111 to 0.1103 relative to uniform routing; it also achieves higher UniMod than hard routing. Because the visual-size bank and downstream configuration are unchanged between uniform and learned-soft routing, these differences provide functional evidence that the diffuse, query-dependent weights produce a useful instance-conditioned prior rather than an effectively class-constant one. The isolated benefit of visual refinement is positive but modest. Removing uncertainty attenuation also reduces UniMod relative to the default configuration.

Prior-strength sensitivity shows no monotonic trend over the tested range. Among the three tested strengths, the $0.5\times$ setting gives the highest median UniMod but also the highest unified MAE and outlier rate. The default setting gives the lowest median unified MAE and the narrowest UniMod interquartile range, while $2\times$ strength gives an intermediate size error and slightly higher median UniMod than the default with greater run-to-run dispersion. Because the UniMod intervals overlap, the results indicate robustness and an AP--size-error trade-off rather than a unique optimum. The CAP-only variant remains competitive in UniMod but has worse unified MAE, supporting CAP as a complementary regulariser rather than a standalone source of size-error reduction.

\begin{table}[htbp]
  \centering
  \begin{singlespace}
  \caption{Size-path diagnostic ablation on KITTI validation. Within each matched run, size-quality metrics are computed on the same ground-truth instances: the intersection of objects matched as true positives by MonoDGP, Injection only, and Injection + CAP. The common set contains a median of 2,286 objects per run (range 2,252--2,458). Values are medians over five runs. UniMod is the detector-level five-run endpoint and is included as supporting context. Best median results in each column are shown in \textbf{bold}.}
  \label{tab:monoprio_mechanism_ablation}
  \normalsize
  \setlength{\tabcolsep}{5.0pt}
  \renewcommand{\arraystretch}{1.15}
  \begin{tabular*}{\textwidth}{@{\extracolsep{\fill}}lccccc@{}}
  \toprule
  Method &
  \makecell[c]{Car\\MAE$\downarrow$} &
  \makecell[c]{Ped.\\MAE$\downarrow$} &
  \makecell[c]{Cyc.\\MAE$\downarrow$} &
  \makecell[c]{Uni\\MAE$\downarrow$} &
  UniMod$\uparrow$ \\
  \midrule
  MonoDGP & \textbf{0.1036} & \textbf{0.0785} & 0.1037 & \textbf{0.1017} & 12.105 \\
  Injection only & 0.1060 & 0.0800 & 0.0928 & 0.1041 & 12.294 \\
  Injection + CAP & 0.1040 & 0.0808 & \textbf{0.0898} & 0.1022 & \textbf{12.307} \\
  \bottomrule
  \end{tabular*}
  \end{singlespace}
\end{table}

\begin{table*}[htbp]
  \centering
  \begin{singlespace}
    \caption{Prior, routing, and prior-strength ablations on KITTI validation. Values are medians [Q1, Q3] over five runs. Uni MAE and Out.@20\% use matched true positives; UniMod averages Moderate AP$_{3D|R40}$ across classes. CM, Size, and VS denote class-mean, size-only, and visual-size banks; routing modes are fixed, uniform, soft, and hard. The final three rows vary only the base injection strength $\lambda_0$, with $1\times$ denoting the default Full MonoPRIO configuration. Inj., Unc., and CAP denote size injection, uncertainty attenuation, and CAP regularisation. Best median results are shown in \textbf{bold}.}
  \label{tab:simple_prior_ablation}
  \scriptsize
  \setlength{\tabcolsep}{3.0pt}
  \renewcommand{\arraystretch}{1.12}
  \resizebox{\textwidth}{!}{%
  \begin{tabular}{llcccccc}
  \toprule
  Variant & Bank/route & Inj. & Unc. & CAP & Uni MAE$\downarrow$ & Out.@20\%$\downarrow$ & UniMod$\uparrow$ \\
  \midrule
  MonoDGP baseline
  & -- & -- & -- & --
  & \makecell[c]{0.1106\\{\scriptsize [0.1105, 0.1108]}}
  & \makecell[c]{0.0011\\{\scriptsize [0.0010, 0.0013]}}
  & \makecell[c]{12.105\\{\scriptsize [11.951, 12.415]}} \\

  Class-mean prior
  & CM/fixed & Yes & Yes & No
  & \makecell[c]{0.1119\\{\scriptsize [0.1117, 0.1128]}}
  & \makecell[c]{\textbf{0.0008}\\{\scriptsize [0.0008, 0.0011]}}
  & \makecell[c]{11.677\\{\scriptsize [11.502, 11.754]}} \\

  Size-only bank
  & Size/uniform & Yes & Yes & Yes
  & \makecell[c]{0.1116\\{\scriptsize [0.1111, 0.1116]}}
  & \makecell[c]{0.0013\\{\scriptsize [0.0011, 0.0016]}}
  & \makecell[c]{11.795\\{\scriptsize [11.327, 11.918]}} \\

  Visual-size uniform
  & VS/uniform & Yes & Yes & Yes
  & \makecell[c]{0.1111\\{\scriptsize [0.1111, 0.1118]}}
  & \makecell[c]{0.0010\\{\scriptsize [0.0010, 0.0016]}}
  & \makecell[c]{11.861\\{\scriptsize [11.620, 12.045]}} \\
  Injection only
  & VS/soft & Yes & Yes & No
  & \makecell[c]{0.1104\\{\scriptsize [0.1100, 0.1112]}}
  & \makecell[c]{\textbf{0.0008}\\{\scriptsize [0.0005, 0.0010]}}
  & \makecell[c]{12.294\\{\scriptsize [11.968, 12.333]}} \\

  CAP only
  & VS/soft & No & -- & Yes
  & \makecell[c]{0.1145\\{\scriptsize [0.1113, 0.1149]}}
  & \makecell[c]{0.0013\\{\scriptsize [0.0008, 0.0015]}}
  & \makecell[c]{12.230\\{\scriptsize [12.077, 12.691]}} \\

  No uncertainty gate
  & VS/soft & Yes & No & Yes
  & \makecell[c]{0.1105\\{\scriptsize [0.1101, 0.1106]}}
  & \makecell[c]{0.0013\\{\scriptsize [0.0005, 0.0020]}}
  & \makecell[c]{12.096\\{\scriptsize [11.972, 12.294]}} \\

  Hard routing
  & VS/hard & Yes & Yes & Yes
  & \makecell[c]{0.1112\\{\scriptsize [0.1110, 0.1125]}}
  & \makecell[c]{0.0013\\{\scriptsize [0.0013, 0.0013]}}
  & \makecell[c]{11.923\\{\scriptsize [11.723, 12.028]}} \\

  \midrule

  $0.5\times\lambda_0$
  & VS/soft & Yes & Yes & Yes
  & \makecell[c]{0.1124\\{\scriptsize [0.1115, 0.1128]}}
  & \makecell[c]{0.0016\\{\scriptsize [0.0008, 0.0016]}}
  & \makecell[c]{\textbf{12.642}\\{\scriptsize [11.890, 12.714]}} \\

  Full MonoPRIO (default)
  & VS/soft & Yes & Yes & Yes
  & \makecell[c]{\textbf{0.1103}\\{\scriptsize [0.1092, 0.1123]}}
  & \makecell[c]{0.0010\\{\scriptsize [0.0008, 0.0013]}}
  & \makecell[c]{12.307\\{\scriptsize [12.250, 12.608]}} \\

  $2\times\lambda_0$
  & VS/soft & Yes & Yes & Yes
  & \makecell[c]{0.1111\\{\scriptsize [0.1104, 0.1112]}}
  & \makecell[c]{0.0010\\{\scriptsize [0.0005, 0.0016]}}
  & \makecell[c]{12.375\\{\scriptsize [11.842, 12.833]}} \\
  \bottomrule
  \end{tabular}}
  \end{singlespace}
\end{table*}

\subsection{Qualitative Results}

\begin{figure}[htbp]
  \centering
  \includegraphics[width=\textwidth,height=0.82\textheight,keepaspectratio]{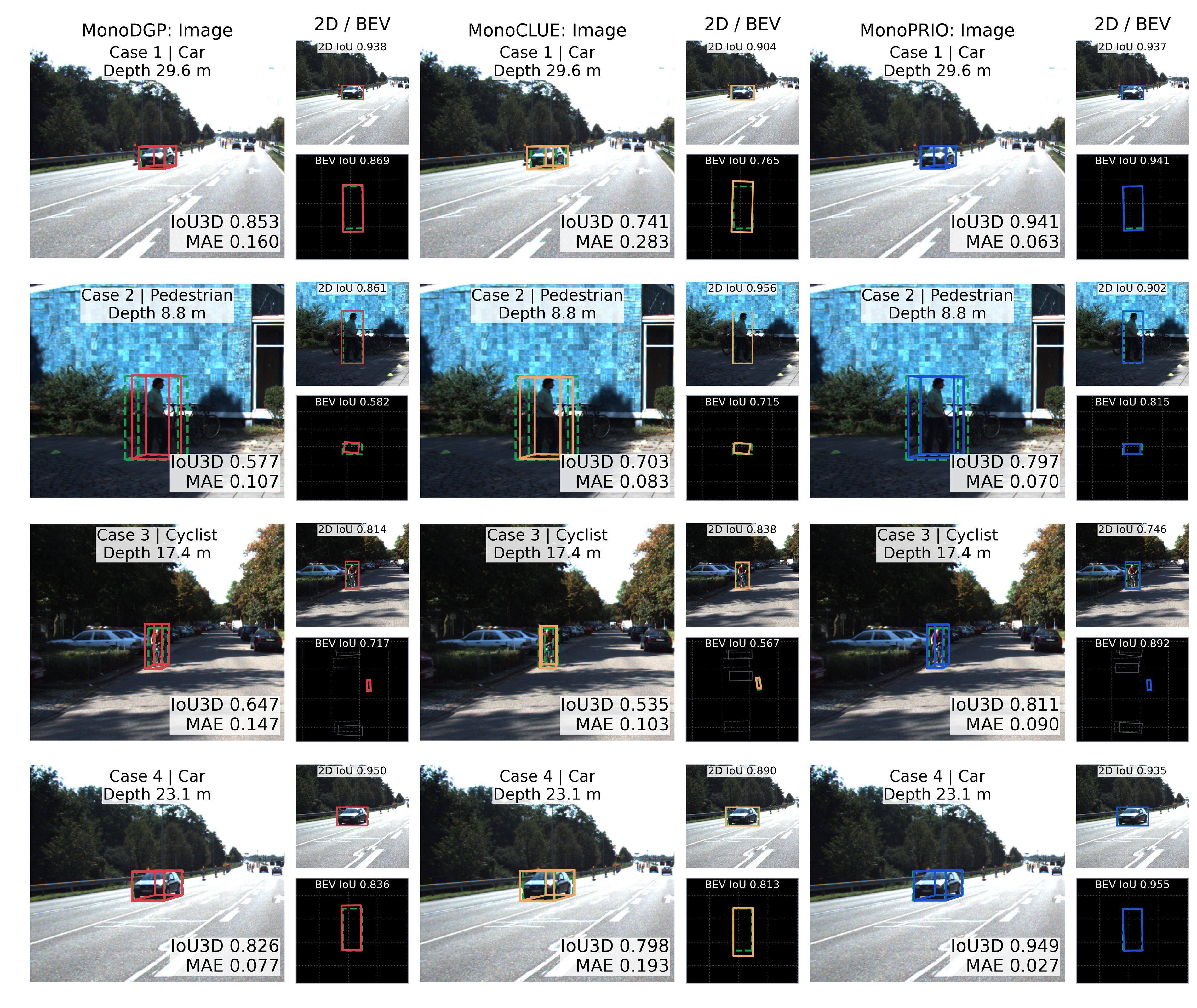}
  \caption{\textbf{Qualitative comparison on KITTI validation.} Representative ambiguity-prone cases with scale uncertainty. Dashed green boxes/footprints denote ground truth; solid red/orange/blue denote MonoDGP/MonoCLUE/MonoPRIO predictions. Insets show 2D overlap and local BEV alignment.}
  \label{fig:monoprio_qualitative}
\end{figure}

Figure~\ref{fig:monoprio_qualitative} provides representative difficult cases. Across these examples, MonoPRIO generally yields tighter localisation and more consistent BEV footprints than the compared baselines. The clearest visual gains appear in partially visible or truncation-prone cases, where compared baselines more often over- or under-estimate object extent.

\subsection{Ambiguity-Regime and Routing Analysis}

To evaluate where adaptive priors are most effective, we analyse matched true positives by occlusion level, routed uncertainty, and prototype-routing concentration. These diagnostics are designed to separate three related effects. Occlusion bins test whether the gains align with degraded visual evidence, routed-uncertainty bins test whether the mixture prior produces a meaningful confidence signal, and prototype-utilisation statistics show how CAP changes the use of the prior bank during training. All three analyses are computed on matched true positives and focus on size-estimation behaviour rather than conflating size errors with false positives or missed detections.

Table~\ref{tab:monoprio_occlusion} shows a selective-gain pattern across visibility regimes. MonoPRIO is slightly behind MonoDGP for fully visible instances, but is best in partly and largely occluded bins. This distribution of gains suggests that adaptive priors are most effective when visual evidence is degraded.

\begin{table}[htbp]
\centering
  \begin{singlespace}
  \caption{Occlusion-stratified unified size error. Values are five-run medians of unified 3D size MAE on matched true positives.}
  \label{tab:monoprio_occlusion}
  \normalsize
  \setlength{\tabcolsep}{6.0pt}
  \renewcommand{\arraystretch}{1.15}
  \begin{tabular*}{\textwidth}{@{\extracolsep{\fill}}lccc@{}}
  \toprule
  Method & Fully visible & Partly occluded & Largely occluded \\
  \midrule
  MonoDGP & \textbf{0.1068} & \underline{0.1091} & \underline{0.1231} \\
  MonoCLUE & 0.1087 & 0.1125 & 0.1281 \\
  MonoPRIO & \underline{0.1081} & \textbf{0.1087} & \textbf{0.1220} \\
  \bottomrule
  \end{tabular*}
  \end{singlespace}
\end{table}

Table~\ref{tab:prior_conf_bins} links routed uncertainty to error behaviour. The high-uncertainty bin generally exhibits the largest size error and outlier rate, supporting $\hat{\sigma}_i$ as a confidence-related indicator for prior assignment. Relative to Injection only, CAP reduces size MAE and outlier ratio in the high-uncertainty bin, but its effect is not uniform across uncertainty regimes. The mid-uncertainty bin metrics worsen, while the low-uncertainty bin shows lower MAE but a higher outlier ratio. CAP is therefore interpreted as providing a selective complementary benefit for higher-uncertainty cases rather than a general improvement across all routed assignments.

\begin{table}[htbp]
  \centering
  \begin{singlespace}
    \caption{Size error versus routed-prior uncertainty on matched KITTI validation true positives. Values are medians over five runs. Uncertainty bins are tertiles of routed uncertainty $\hat{\sigma}_i$, where lower uncertainty indicates higher-confidence prior assignment. Size MAE denotes mean absolute size error, Rel MAE denotes mean relative size error averaged over $(h,w,l)$, and Outlier ratio@20\% denotes the fraction of matched true positives whose mean relative size error exceeds 20\%.}
  \label{tab:prior_conf_bins}
  \normalsize
  \setlength{\tabcolsep}{5.0pt}
  \renewcommand{\arraystretch}{1.15}
  \begin{tabular*}{\textwidth}{@{\extracolsep{\fill}}llccc@{}}
  \toprule
  Method & \makecell[c]{Uncertainty\\bin} & \makecell[c]{Size\\MAE$\downarrow$} & \makecell[c]{Rel\\MAE$\downarrow$} & \makecell[c]{Outlier ratio\\@20\%$\downarrow$} \\
  \midrule
  Injection + CAP & Low  & 0.1095 & 0.0469 & 0.0009 \\
  Injection only  & Low  & 0.1112 & 0.0472 & 0.0003 \\
  Injection + CAP & Mid  & 0.1110 & 0.0459 & 0.0008 \\
  Injection only  & Mid  & 0.1082 & 0.0444 & 0.0005 \\
  Injection + CAP & High & 0.1117 & 0.0519 & 0.0012 \\
  Injection only  & High & 0.1130 & 0.0519 & 0.0016 \\
  \bottomrule
  \end{tabular*}
  \end{singlespace}
\end{table}

Table~\ref{tab:prototype_utilization} characterises how CAP changes routing behaviour on matched KITTI validation true positives. Mean maximum weight and normalised entropy describe the softness of the bank-wide routed mixture, while Top-1 usage and active prototypes describe the concentration of largest-weight prototype indices. CAP does not make individual routing distributions sharply peaked: mean maximum weights remain around 0.07--0.10 and normalised entropies remain high. It primarily changes which prototype attains the largest routing weight, producing a pronounced concentration of prototype-index usage for Pedestrian and Cyclist. Because this concentration may reflect either more consistent routing or reduced prototype diversity, it is not interpreted as beneficial in isolation and is considered only alongside the corresponding size-error and AP diagnostics.

For matched query $i$, normalised routing entropy is computed as

\begin{equation}
H_i =
-\frac{\sum_{k=1}^{K} a_{ik}\log a_{ik}}{\log K},
\end{equation}

where $K=35$ is the complete unified prototype bank. Diagnostics are grouped by ground-truth class, but routing maxima and argmax indices are computed over the complete bank.

\begin{table}[htbp]
\centering
  \begin{singlespace}
  \caption{Prototype-routing diagnostics on matched KITTI validation true positives, reported as medians over five runs. Statistics are grouped by ground-truth class but computed over the complete 35-prototype routing distribution. Mean maximum weight and normalised entropy describe soft routing sharpness. Top-1 usage is the fraction of samples whose bank-wide largest-weight index equals the most frequently occurring largest-weight index for that class group. Active prototype count denotes the number of distinct bank-wide largest-weight indices.}
  \label{tab:prototype_utilization}
  \scriptsize
  \setlength{\tabcolsep}{3.2pt}
  \renewcommand{\arraystretch}{1.14}
  \begin{tabular}{@{}llccc@{}}
  \toprule
  Diagnostic & Class &
  \makecell[c]{Injection\\only} &
  \makecell[c]{Injection\\+ CAP} &
  $\Delta$ \\
  \midrule
  \multirow{3}{*}{Mean max}
  & Car & 0.1008 & 0.1003 & -0.0005 \\
  & Pedestrian & 0.0711 & 0.0737 & +0.0026 \\
  & Cyclist & 0.0685 & 0.0699 & +0.0014 \\
  \midrule
  \multirow{3}{*}{Norm. entropy}
  & Car & 0.7089 & 0.7136 & +0.0047 \\
  & Pedestrian & 0.8189 & 0.8138 & -0.0051 \\
  & Cyclist & 0.8512 & 0.8571 & +0.0059 \\
  \midrule
  \multirow{3}{*}{Top-1 usage}
  & Car & 0.3257 & 0.3767 & +0.0510 \\
  & Pedestrian & 0.4201 & 0.9965 & +0.5764 \\
  & Cyclist & 0.3086 & 0.8987 & +0.5901 \\
  \midrule
  \multirow{3}{*}{\makecell[l]{Active prototypes\\(of 35)}}
  & Car & 10 & 5 & -5 \\
  & Pedestrian & 12 & 2 & -10 \\
  & Cyclist & 7 & 4 & -3 \\
  \bottomrule
  \end{tabular}
  \end{singlespace}
\end{table}

\subsection{Low-Data Behaviour}

Table~\ref{tab:lowdata_prior_strength_variant} reports the results of the 20\% and 40\% experiments. For each split, the prior bank is rebuilt solely from the corresponding training subset using the same deterministic construction procedure and soft class-aware routing. The 20\% and 40\% banks contain $[9,5,8]$ and $[10,6,9]$ prototypes, respectively, in Pedestrian, Car, and Cyclist order. All methods use the same data fractions and five-run summaries, with the same 250-epoch training budget, learning rate, weight decay, and step-decay schedule. Within each method and split, the recorded configuration is fixed across the five seeds, but we do not claim that every method received an exhaustive matched hyperparameter search.

With both 20\% and 40\% of the training data, MonoPRIO has the highest median in all reported categories, with the largest margins on Cyclist and Pedestrian. These results show that the tested MonoPRIO configuration remains effective when supervision is limited. Because the study did not use an exhaustive matched hyperparameter search for every method, we interpret Table~\ref{tab:lowdata_prior_strength_variant} as a robustness comparison under the reported configurations rather than evidence of globally optimal low-data tuning.

\begin{table}[htbp]
\centering
\begin{singlespace}
  \caption{Low-data comparison at 20\% and 40\% data under the reported configurations. Values are AP$_{3D|R40}$ Moderate medians over five runs. Car uses IoU=0.7. Pedestrian/Cyclist use IoU=0.5. UniMod is additionally reported as median [Q1, Q3] over the same five runs to show run-to-run dispersion. $\Delta$ is MonoPRIO minus the second-best median in each column. Best and second-best medians are shown in \textbf{bold} and \underline{underlined}, respectively.}
  \label{tab:lowdata_prior_strength_variant}
  \normalsize
  \setlength{\tabcolsep}{8.0pt}
  \renewcommand{\arraystretch}{1.18}
  \begin{tabular*}{0.9\textwidth}{@{\extracolsep{\fill}}llcccc@{}}
    \toprule
    Data split & Method & Car & Ped. & Cyc. & UniMod \\
    \midrule
    20\%
      & MonoDGP  & 15.446 & 1.426 & \underline{2.032} & \makecell[c]{6.281\\{\scriptsize [6.248, 6.697]}} \\
      & MonoCLUE & \underline{15.936} & \underline{2.141} & 1.864 & \makecell[c]{\underline{6.667}\\{\scriptsize [6.657, 6.731]}} \\
      & MonoPRIO & \textbf{16.086} & \textbf{3.455} & \textbf{4.700} & \makecell[c]{\textbf{8.154}\\{\scriptsize [7.838, 8.535]}} \\
      & $\Delta$ & +0.150 & +1.314 & +2.668 & +1.487 \\
    \midrule
    40\%
      & MonoDGP  & \underline{18.798} & \underline{5.130} & \underline{3.916} & \makecell[c]{\underline{9.348}\\{\scriptsize [9.335, 9.528]}} \\
      & MonoCLUE & 18.186 & 4.513 & 3.413 & \makecell[c]{9.028\\{\scriptsize [6.635, 9.190]}} \\
      & MonoPRIO & \textbf{19.355} & \textbf{5.860} & \textbf{6.301} & \makecell[c]{\textbf{10.131}\\{\scriptsize [10.128, 10.669]}} \\
      & $\Delta$ & +0.557 & +0.730 & +2.385 & +0.783 \\
    \bottomrule
  \end{tabular*}
\end{singlespace}
\end{table}

\subsection{Failure Cases}

\begin{figure}[!htbp]
  \centering
  \includegraphics[width=\textwidth,height=0.82\textheight,keepaspectratio]{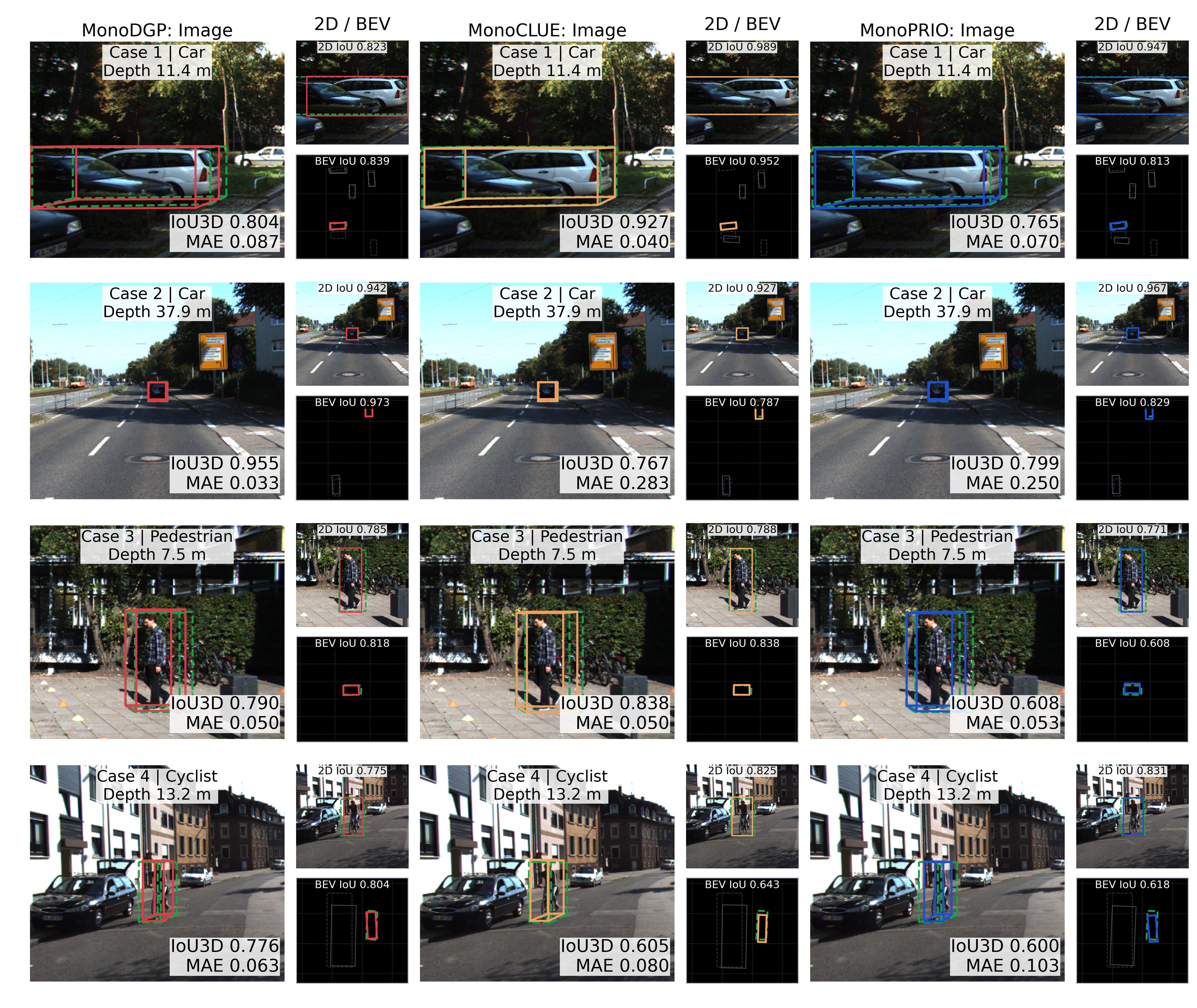}
  \caption{\textbf{Failure cases on KITTI validation.} Representative low-IoU cases where MonoPRIO underperforms, covering severe occlusion, scale-depth ambiguity, and prior-geometry mismatch. Insets show 2D overlap and local BEV alignment.}
  \label{fig:monoprio_qualitative_failures}
\end{figure}

Figure~\ref{fig:monoprio_qualitative_failures} highlights boundary conditions where MonoPRIO can underperform. Severe occlusion leaves limited visual evidence, which can destabilise routing and degrade local 3D fit and BEV alignment. Long-range scale-depth ambiguity arises from very small image extent at large distances, where multiple metric interpretations remain plausible and prior conditioning can select an inappropriate mode. The largely visible pedestrian is consistent with a prior-geometry mismatch, in which prior conditioning biases size and localisation toward a plausible but unsuitable mode. Together, these cases suggest that adaptive priors are most effective when ambiguity is structured and the bank adequately represents the observed instance, and are less reliable when evidence is limited or the instance lies beyond the support estimated from the training distribution.

To quantify this boundary condition, we define a support distance for each validation object as the nearest standardised log-size distance from its ground-truth dimensions to a same-class prototype manifold:

\begin{equation}
  s_i =
  \min_{k\in\mathcal{K}_{c_i}}
  \left\|
  \left(
  V_k^{\log\top}
  \left(\log d_i-\mu_k^{\log}\right)
  \right)
  \odot
  \eta_k^{-1/2}
  \right\|_2,
  \label{eq:support_distance}
\end{equation}

where $d_i=(h_i,w_i,l_i)$ and $\mathcal{K}_{c_i}$ denotes the prototypes belonging to the ground-truth class. We define a class-specific support threshold using only the retained training instances used to construct the corresponding prior bank:

\begin{equation}
  \tau_c = Q_{0.95}\left(\left\{s_j^{\mathrm{train}}:y_j=c\right\}\right),
  \label{eq:support_threshold}
\end{equation}

where $Q_{0.95}$ denotes the 95th percentile. A validation object is classified as in-support when $s_i\leq\tau_{c_i}$ and out-of-support otherwise. The thresholds are therefore fixed independently of the validation results. The resulting thresholds are $\tau_{\mathrm{Ped}}=1.9352$, $\tau_{\mathrm{Car}}=1.2436$, and $\tau_{\mathrm{Cyc}}=1.4095$. Because this diagnostic uses ground-truth class labels and dimensions, it is used only for analysis and is not available as an inference-time out-of-support detector.

\begin{table}[t]
  \centering
    \caption{Training-derived out-of-support analysis on KITTI validation. Support thresholds are class-specific training-set 95th percentiles. GT gives count and percentage; Match is the five-run median fraction of ground-truth objects matched at class-specific 3D-IoU thresholds of 0.7 for Car and 0.5 for Pedestrian/Cyclist; other metrics are five-run medians over matched true positives.}
  \label{tab:out_of_support}
  \small
  \setlength{\tabcolsep}{3.5pt}
  \renewcommand{\arraystretch}{1.12}
  \begin{tabular*}{\textwidth}{@{\extracolsep{\fill}}lcccccc@{}}
  \toprule
  \makecell[l]{Support\\status} &
  GT (\%) &
  Match$\uparrow$ &
  MAE$\downarrow$ &
  Rel.$\downarrow$ &
  \makecell[c]{Out.\\@20\%$\downarrow$} &
  \makecell[c]{mIoU$_{\mathrm{3D}}$\\$\uparrow$} \\
  \midrule
  In-support & 15,736 (89.6) & 0.2359 & 0.1074 & 0.0466 & 0.0006 & 0.7762 \\
  Out-of-support & 1,822 (10.4) & 0.0922 & 0.1687 & 0.0837 & 0.0106 & 0.7271 \\
  \bottomrule
  \end{tabular*}
\end{table}

Table~\ref{tab:out_of_support} quantifies performance beyond the support boundary defined from the training distribution. Out-of-support objects account for 10.4\% of the 17,558 target-class objects in the KITTI validation split and have a substantially lower median match rate than in-support objects (0.0922 versus 0.2359). Among matched true positives, out-of-support objects have higher size MAE (0.1687 versus 0.1074), higher relative size MAE (0.0837 versus 0.0466), and lower median matched mIoU$_{\mathrm{3D}}$ (0.7271 versus 0.7762). Their median Outlier ratio@20\% is also higher (0.0106 versus 0.0006), although this statistic is based on a small number of large-error cases per run. These results show that objects beyond the training-derived prototype-support threshold are substantially harder for MonoPRIO. The analysis does not establish that prior conditioning causes every failure in this group, but it identifies atypical size geometry as a measurable boundary condition for the method.

\subsection{When Adaptive Priors Help, and When They Hurt}

Across the unified KITTI setting, MonoPRIO provides its clearest benefits when monocular evidence is ambiguous, particularly under partial occlusion and data-scarce training, while gains are smaller or less consistent for fully visible objects and in the less ambiguous car-only setting (Tables~\ref{tab:monoprio_occlusion}, \ref{tab:lowdata_prior_strength_variant}, and~\ref{tab:car_test_val_monoprio}). The mechanism analyses show different roles for the two components. Routed injection provides the clearest reduction in large size failures, while CAP shows selective class and uncertainty-dependent effects rather than a consistent aggregate UniMod improvement. Changes in routing concentration are likewise treated as descriptive evidence of altered prototype use rather than as evidence of improved performance by themselves.

The effectiveness of adaptive conditioning therefore depends on both prior-bank coverage and reliable query routing. The failure cases and training-derived support analysis show that atypical geometries and severe visibility loss can still produce a mismatch between the routed prior and the true object dimensions (Figure~\ref{fig:monoprio_qualitative_failures} and Table~\ref{tab:out_of_support}). MonoPRIO should consequently be viewed as a branch-local stabilisation mechanism for structured size ambiguity rather than as a universal correction for monocular 3D errors. Future work should focus on broader prototype coverage, explicit detection of out-of-support assignments, and stronger adaptation of prior influence to visibility, uncertainty, and routing confidence.

\section{Conclusion}

MonoPRIO introduces branch-local adaptive prior conditioning for metric size prediction in a MonoDGP-style monocular detector, combining an offline class-aware prototype bank, soft query routing, uncertainty-aware log-space injection, and CAP regularisation. On KITTI, it improves unified detection while retaining near-baseline efficiency, and the nuScenes frontal evaluation shows that the mechanism remains beneficial when the bank is rebuilt for a second dataset. Ablations indicate that routed injection provides the main reduction in large size failures, while CAP provides selective complementary effects. Performance nevertheless depends on representative prior-bank coverage and reliable routing, particularly under atypical geometry or severe occlusion. Future work should address broader prototype coverage, out-of-support detection, and adaptive control of prior strength.

\section*{CRediT authorship contribution statement}

Leon Davies: Conceptualization, Methodology, Software, Validation, Formal analysis, Investigation, Data curation, Visualization, Writing -- original draft, Writing -- review and editing. Qinggang Meng: Supervision, Conceptualization, Methodology, Funding acquisition, Resources, Writing -- review and editing. Mohamad Saada: Writing -- review and editing. Baihua Li: Supervision, Writing -- review and editing. Simon S{\o}lvsten: Supervision, Project administration, Writing -- review and editing.

\section*{Acknowledgements}

This research was supported by the Willis Research Network (WRN), part of WTW. No grant number was associated with this support. The funder had no role in the design of the study, analysis and interpretation of results, writing of the manuscript, or decision to submit the article for publication. The authors also acknowledge the use of the Apollo High Performance Computing service at Loughborough University.

\bibliographystyle{elsarticle-num}
\bibliography{references}

@inproceedings{geiger2012kitti,
  author    = {Geiger, Andreas and Lenz, Philip and Urtasun, Raquel},
  title     = {Are We Ready for Autonomous Driving? The {KITTI} Vision Benchmark Suite},
  booktitle = {Proceedings of the IEEE Conference on Computer Vision and Pattern Recognition (CVPR)},
  pages     = {3354--3361},
  year      = {2012},
  doi       = {10.1109/CVPR.2012.6248074}
}

@article{dong2026monoa2,
  author  = {Dong, Jinpeng and Zhou, Sanping and Hu, Yufeng and Huang, Yuhao and Jiang, Jingjing and Zuo, Weiliang and Chen, Shitao and Zheng, Nanning},
  title   = {{MonoA}$^2$: Adaptive Depth with Augmented Head for Monocular {3D} Object Detection},
  journal = {Pattern Recognition},
  volume  = {172},
  pages   = {112418},
  year    = {2026},
  doi     = {10.1016/j.patcog.2025.112418}
}

@inproceedings{radford2021clip,
  author    = {Radford, Alec and Kim, Jong Wook and Hallacy, Chris and Ramesh, Aditya and Goh, Gabriel and Agarwal, Sandhini and Sastry, Girish and Askell, Amanda and Mishkin, Pamela and Clark, Jack and Krueger, Gretchen and Sutskever, Ilya},
  title     = {Learning Transferable Visual Models From Natural Language Supervision},
  booktitle = {Proceedings of the 38th International Conference on Machine Learning (ICML)},
  series    = {Proceedings of Machine Learning Research},
  volume    = {139},
  pages     = {8748--8763},
  year      = {2021},
  publisher = {PMLR}
}

@inproceedings{lu2021gupnet,
  author    = {Lu, Yan and Ma, Xinzhu and Yang, Lei and Zhang, Tianzhu and Liu, Yating and Chu, Qi and Yan, Junjie and Ouyang, Wanli},
  title     = {Geometry Uncertainty Projection Network for Monocular {3D} Object Detection},
  booktitle = {Proceedings of the IEEE/CVF International Conference on Computer Vision (ICCV)},
  pages     = {3111--3121},
  year      = {2021}
}

@inproceedings{kumar2022deviant,
  author    = {Kumar, Abhinav and Brazil, Garrick and Corona, Enrique and Parchami, Armin and Liu, Xiaoming},
  title     = {{DEVIANT}: Depth EquiVarIAnt NeTwork for Monocular {3D} Object Detection},
  booktitle = {Computer Vision -- ECCV 2022},
  series    = {Lecture Notes in Computer Science},
  volume    = {13669},
  pages     = {664--683},
  year      = {2022},
  doi       = {10.1007/978-3-031-20077-9_39}
}

@article{liu2022monocon,
  author    = {Liu, Xianpeng and Xue, Nan and Wu, Tianfu},
  title     = {Learning Auxiliary Monocular Contexts Helps Monocular {3D} Object Detection},
  journal   = {Proceedings of the AAAI Conference on Artificial Intelligence},
  volume    = {36},
  number    = {2},
  pages     = {1810--1818},
  year      = {2022},
  doi       = {10.1609/aaai.v36i2.20074}
}

@inproceedings{jia2023monouni,
  author    = {Jia, Jinrang and Li, Zhenjia and Shi, Yifeng},
  title     = {{MonoUNI}: A Unified Vehicle and Infrastructure-Side Monocular {3D} Object Detection Network with Sufficient Depth Clues},
  booktitle = {Advances in Neural Information Processing Systems},
  volume    = {36},
  year      = {2023}
}

@article{su2023opa3d,
  author  = {Su, Yongzhi and Di, Yan and Zhai, Guangyao and Manhardt, Fabian and Rambach, Jason R. and Busam, Benjamin and Stricker, Didier and Tombari, Federico},
  title   = {{OPA-3D}: Occlusion-Aware Pixel-Wise Aggregation for Monocular {3D} Object Detection},
  journal = {IEEE Robot. Autom. Lett.},
  volume  = {8},
  number  = {3},
  pages   = {1327--1334},
  year    = {2023},
  doi     = {10.1109/LRA.2023.3238137}
}

@inproceedings{li2022monodde,
  author    = {Li, Zhuoling and Qu, Zhan and Zhou, Yang and Liu, Jianzhuang and Wang, Haoqian and Jiang, Lihui},
  title     = {Diversity Matters: Fully Exploiting Depth Clues for Reliable Monocular {3D} Object Detection},
  booktitle = {Proceedings of the IEEE/CVF Conference on Computer Vision and Pattern Recognition (CVPR)},
  pages     = {2791--2800},
  year      = {2022}
}

@article{guan2022monopoly,
  author  = {Guan, He and Song, Chunfeng and Zhang, Zhaoxiang and Tan, Tieniu},
  title   = {{MonoPoly}: A Practical Monocular {3D} Object Detector},
  journal = {Pattern Recognition},
  volume  = {132},
  pages   = {108967},
  year    = {2022},
  doi     = {10.1016/j.patcog.2022.108967}
}

@inproceedings{pu2025monodgp,
  author    = {Pu, Fanqi and Wang, Yifan and Deng, Jiru and Yang, Wenming},
  title     = {{MonoDGP}: Monocular {3D} Object Detection with Decoupled-Query and Geometry-Error Priors},
  booktitle = {Proceedings of the IEEE/CVF Conference on Computer Vision and Pattern Recognition (CVPR)},
  pages     = {6520--6530},
  year      = {2025}
}

@inproceedings{zhang2025monocop,
  author    = {Zhang, Zhihao and Kumar, Abhinav and Ganesan, Girish Chandar and Liu, Xiaoming},
  title     = {Unleashing the Power of Chain-of-Prediction for Monocular {3D} Object Detection},
  booktitle = {Proceedings of the IEEE/CVF Conference on Computer Vision and Pattern Recognition (CVPR)},
  pages     = {18755--18765},
  year      = {2026}
}

@article{yang2025monoclue,
  author    = {Yang, Sunghun and Lee, Minhyeok and Lee, Jungho and Lee, Sangyoun},
  title     = {{MonoCLUE}: Object-Aware Clustering Enhances Monocular {3D} Object Detection},
  journal   = {Proceedings of the AAAI Conference on Artificial Intelligence},
  volume    = {40},
  number    = {14},
  pages     = {11721--11729},
  year      = {2026},
  doi       = {10.1609/aaai.v40i14.38157}
}

@inproceedings{zhang2023monodetr,
  author    = {Zhang, Renrui and Qiu, Han and Wang, Tai and Guo, Ziyu and Cui, Ziteng and Qiao, Yu and Li, Hongsheng and Gao, Peng},
  title     = {{MonoDETR}: Depth-Guided Transformer for Monocular {3D} Object Detection},
  booktitle = {Proceedings of the IEEE/CVF International Conference on Computer Vision (ICCV)},
  pages     = {9155--9166},
  year      = {2023}
}

@inproceedings{wu2023monopgc,
  author    = {Wu, Zizhang and Gan, Yuanzhu and Wang, Lei and Chen, Guilian and Pu, Jian},
  title     = {{MonoPGC}: Monocular {3D} Object Detection with Pixel Geometry Contexts},
  booktitle = {Proceedings of the IEEE International Conference on Robotics and Automation (ICRA)},
  pages     = {4842--4849},
  year      = {2023},
  doi       = {10.1109/ICRA48891.2023.10161442}
}

@inproceedings{huang2022monodtr,
  author    = {Huang, Kuan-Chih and Wu, Tsung-Han and Su, Hung-Ting and Hsu, Winston H.},
  title     = {{MonoDTR}: Monocular {3D} Object Detection with Depth-Aware Transformer},
  booktitle = {Proceedings of the IEEE/CVF Conference on Computer Vision and Pattern Recognition (CVPR)},
  pages     = {4012--4021},
  year      = {2022}
}

@inproceedings{peng2022didm3d,
  author    = {Peng, Liang and Wu, Xiaopei and Yang, Zheng and Liu, Haifeng and Cai, Deng},
  title     = {{DID-M3D}: Decoupling Instance Depth for Monocular {3D} Object Detection},
  booktitle = {Computer Vision -- ECCV 2022},
  series    = {Lecture Notes in Computer Science},
  volume    = {13661},
  pages     = {71--88},
  year      = {2022},
  doi       = {10.1007/978-3-031-19769-7_5}
}

@inproceedings{peng2024occupancym3d,
  author    = {Peng, Liang and Xu, Junkai and Cheng, Haoran and Yang, Zheng and Wu, Xiaopei and Qian, Wei and Wang, Wenxiao and Wu, Boxi and Cai, Deng},
  title     = {Learning Occupancy for Monocular {3D} Object Detection},
  booktitle = {Proceedings of the IEEE/CVF Conference on Computer Vision and Pattern Recognition (CVPR)},
  pages     = {10281--10292},
  year      = {2024}
}

@article{wu2024fd3d,
  author    = {Wu, Zizhang and Gan, Yuanzhu and Wu, Yunzhe and Wang, Ruihao and Wang, Xiaoquan and Pu, Jian},
  title     = {{FD3D}: Exploiting Foreground Depth Map for Feature-Supervised Monocular {3D} Object Detection},
  journal   = {Proceedings of the AAAI Conference on Artificial Intelligence},
  volume    = {38},
  number    = {6},
  pages     = {6189--6197},
  year      = {2024},
  doi       = {10.1609/aaai.v38i6.28436}
}

@inproceedings{jiang2024monomae,
  author    = {Jiang, Xueying and Jin, Sheng and Zhang, Xiaoqin and Shao, Ling and Lu, Shijian},
  title     = {{MonoMAE}: Enhancing Monocular {3D} Detection Through Depth-Aware Masked Autoencoders},
  booktitle = {Advances in Neural Information Processing Systems},
  volume    = {37},
  year      = {2024}
}

@inproceedings{yan2024monocd,
  author    = {Yan, Longfei and Yan, Pei and Xiong, Shengzhou and Xiang, Xuanyu and Tan, Yihua},
  title     = {{MonoCD}: Monocular {3D} Object Detection with Complementary Depths},
  booktitle = {Proceedings of the IEEE/CVF Conference on Computer Vision and Pattern Recognition (CVPR)},
  pages     = {10248--10257},
  year      = {2024}
}

@inproceedings{wang2022detr3d,
  author    = {Wang, Yue and Guizilini, Vitor Campagnolo and Zhang, Tianyuan and Wang, Yilun and Zhao, Hang and Solomon, Justin},
  title     = {{DETR3D}: {3D} Object Detection from Multi-View Images via {3D}-to-{2D} Queries},
  booktitle = {Proceedings of the 5th Conference on Robot Learning (CoRL)},
  series    = {Proceedings of Machine Learning Research},
  volume    = {164},
  pages     = {180--191},
  year      = {2022},
  publisher = {PMLR}
}

@inproceedings{li2022bevformerlearningbirdseyeviewrepresentation,
  author    = {Li, Zhiqi and Wang, Wenhai and Li, Hongyang and Xie, Enze and Sima, Chonghao and Lu, Tong and Qiao, Yu and Dai, Jifeng},
  title     = {{BEVFormer}: Learning Bird's-Eye-View Representation from Multi-Camera Images via Spatiotemporal Transformers},
  booktitle = {Computer Vision -- ECCV 2022},
  series    = {Lecture Notes in Computer Science},
  volume    = {13669},
  pages     = {1--18},
  year      = {2022},
  doi       = {10.1007/978-3-031-20077-9_1}
}

@inproceedings{Chen_2016_CVPR,
  author    = {Chen, Xiaozhi and Kundu, Kaustav and Zhang, Ziyu and Ma, Huimin and Fidler, Sanja and Urtasun, Raquel},
  title     = {Monocular {3D} Object Detection for Autonomous Driving},
  booktitle = {Proceedings of the IEEE Conference on Computer Vision and Pattern Recognition (CVPR)},
  pages     = {2147--2156},
  year      = {2016}
}

@inproceedings{Brazil_2019_ICCV,
  author    = {Brazil, Garrick and Liu, Xiaoming},
  title     = {{M3D-RPN}: Monocular {3D} Region Proposal Network for Object Detection},
  booktitle = {Proceedings of the IEEE/CVF International Conference on Computer Vision (ICCV)},
  pages     = {9287--9296},
  year      = {2019}
}

@article{Qian_2022_PR_Survey,
  author  = {Qian, Rui and Lai, Xin and Li, Xirong},
  title   = {{3D} Object Detection for Autonomous Driving: A Survey},
  journal = {Pattern Recognition},
  volume  = {130},
  pages   = {108796},
  year    = {2022},
  doi     = {10.1016/j.patcog.2022.108796}
}

@inproceedings{liu2025monotakd,
  author    = {Liu, Hou-I and Wu, Christine and Cheng, Jen-Hao and Chai, Wenhao and Wang, Shian-Yun and Liu, Gaowen and Latapie, Hugo and Wu, Jhih-Ciang and Hwang, Jenq-Neng and Shuai, Hong-Han and Cheng, Wen-Huang},
  title     = {{MonoTAKD}: Teaching Assistant Knowledge Distillation for Monocular {3D} Object Detection},
  booktitle = {Proceedings of the IEEE/CVF Conference on Computer Vision and Pattern Recognition (CVPR)},
  pages     = {22266--22275},
  year      = {2025}
}

@inproceedings{zhang2021objects,
  author    = {Zhang, Yunpeng and Lu, Jiwen and Zhou, Jie},
  title     = {Objects Are Different: Flexible Monocular {3D} Object Detection},
  booktitle = {Proceedings of the IEEE/CVF Conference on Computer Vision and Pattern Recognition (CVPR)},
  pages     = {3289--3298},
  year      = {2021}
}

@inproceedings{shi2021geometry,
  author    = {Shi, Xuepeng and Ye, Qi and Chen, Xiaozhi and Chen, Chuangrong and Chen, Zhixiang and Kim, Tae-Kyun},
  title     = {Geometry-Based Distance Decomposition for Monocular {3D} Object Detection},
  booktitle = {Proceedings of the IEEE/CVF International Conference on Computer Vision (ICCV)},
  pages     = {15172--15181},
  year      = {2021}
}

@inproceedings{NIPS2015_6da37dd3,
  author    = {Chen, Xiaozhi and Kundu, Kaustav and Zhu, Yukun and Berneshawi, Andrew G. and Ma, Huimin and Fidler, Sanja and Urtasun, Raquel},
  title     = {{3D} Object Proposals for Accurate Object Class Detection},
  booktitle = {Advances in Neural Information Processing Systems},
  volume    = {28},
  pages     = {424--432},
  year      = {2015}
}

@inproceedings{ding2020learning,
  author    = {Ding, Mingyu and Huo, Yuqi and Yi, Hongwei and Wang, Zhe and Shi, Jianping and Lu, Zhiwu and Luo, Ping},
  title     = {Learning Depth-Guided Convolutions for Monocular {3D} Object Detection},
  booktitle = {Proceedings of the IEEE/CVF Conference on Computer Vision and Pattern Recognition (CVPR)},
  pages     = {11672--11681},
  year      = {2020}
}

@inproceedings{liu2020smoke,
  author    = {Liu, Zechen and Wu, Zizhang and T{\'o}th, Roland},
  title     = {{SMOKE}: Single-Stage Monocular {3D} Object Detection via Keypoint Estimation},
  booktitle = {Proceedings of the IEEE/CVF Conference on Computer Vision and Pattern Recognition (CVPR) Workshops},
  pages     = {996--997},
  year      = {2020}
}

@inproceedings{reading2021categorical,
  author    = {Reading, Cody and Harakeh, Ali and Chae, Julia and Waslander, Steven L.},
  title     = {Categorical Depth Distribution Network for Monocular {3D} Object Detection},
  booktitle = {Proceedings of the IEEE/CVF Conference on Computer Vision and Pattern Recognition (CVPR)},
  pages     = {8555--8564},
  year      = {2021},
  doi       = {10.1109/CVPR46437.2021.00845}
}

@inproceedings{roddick2018orthographic,
  author    = {Roddick, Thomas and Kendall, Alex and Cipolla, Roberto},
  title     = {Orthographic Feature Transform for Monocular {3D} Object Detection},
  booktitle = {Proceedings of the British Machine Vision Conference (BMVC)},
  year      = {2019},
  doi       = {10.5244/C.33.59}
}

@inproceedings{mousavian20173d,
  author    = {Mousavian, Arsalan and Anguelov, Dragomir and Flynn, John and Kosecka, Jana},
  title     = {{3D} Bounding Box Estimation Using Deep Learning and Geometry},
  booktitle = {Proceedings of the IEEE Conference on Computer Vision and Pattern Recognition (CVPR)},
  pages     = {7074--7082},
  year      = {2017}
}

@article{li2023bevdepth,
  author    = {Li, Yinhao and Ge, Zheng and Yu, Guanyi and Yang, Jinrong and Wang, Zengran and Shi, Yukang and Sun, Jianjian and Li, Zeming},
  title     = {{BEVDepth}: Acquisition of Reliable Depth for Multi-View {3D} Object Detection},
  journal   = {Proceedings of the AAAI Conference on Artificial Intelligence},
  volume    = {37},
  number    = {2},
  pages     = {1477--1485},
  year      = {2023},
  doi       = {10.1609/aaai.v37i2.25233}
}

@inproceedings{yang2023bevheight,
  author    = {Yang, Lei and Yu, Kaicheng and Tang, Tao and Li, Jun and Yuan, Kun and Wang, Li and Zhang, Xinyu and Chen, Peng},
  title     = {{BEVHeight}: A Robust Framework for Vision-Based Roadside {3D} Object Detection},
  booktitle = {Proceedings of the IEEE/CVF Conference on Computer Vision and Pattern Recognition (CVPR)},
  pages     = {21611--21620},
  year      = {2023}
}

@article{yang2025bevheightpp,
  author  = {Yang, Lei and Tang, Tao and Li, Jun and Yuan, Kun and Wu, Kai and Chen, Peng and Wang, Li and Huang, Yi and Li, Lei and Zhang, Xinyu and Yu, Kaicheng},
  title   = {{BEVHeight++}: Toward Robust Visual Centric {3D} Object Detection},
  journal = {IEEE Trans. Pattern Anal. Mach. Intell.},
  volume  = {47},
  number  = {6},
  pages   = {5094--5111},
  year    = {2025},
  doi     = {10.1109/TPAMI.2025.3549711}
}

\end{document}